%% file: main.tex
\documentclass{article}

\usepackage[eandd, final]{neurips_2026}

\makeatletter
\renewcommand{\@notice}{}
\makeatother

\setcitestyle{numbers,square,sort&compress}

\usepackage[utf8]{inputenc}
\usepackage[T1]{fontenc}
\usepackage{hyperref}
\usepackage{url}
\usepackage{booktabs}
\usepackage{amsfonts}
\usepackage{amsmath}
\usepackage{amssymb}
\usepackage{nicefrac}
\usepackage{microtype}
\usepackage{xcolor}
\usepackage{graphicx}
\usepackage{tabularx}
\usepackage{makecell}
\usepackage{multirow}
\usepackage{array}
\usepackage{longtable}
\usepackage{pdflscape}
\usepackage{enumitem}

\title{EndoSCOP-V: A Multi-Turn Video Understanding Evaluation Framework
for Multimodal Models in Endoscopy Reporting and Clinical Reasoning}

\author{%
  Abhishek Tyagi\textsuperscript{1,*}, Tanmay Jain\textsuperscript{2}, Vilas Biradar\textsuperscript{2}, Krithi Krishna Koduri\textsuperscript{2},\\
  \bf Misba Hul Bashir Masoodi\textsuperscript{2}, Swetha Rapolu\textsuperscript{2}, Hardik Rughwani\textsuperscript{2}, Rachana Kotha\textsuperscript{1},\\
  \bf Anjan Kumar Kaipa\textsuperscript{2}, Anudeep Katrevula\textsuperscript{2}, Rakesh Garlapati\textsuperscript{2}, Aniruddha Pratap Singh\textsuperscript{2},\\
  \bf Rajendra Patel\textsuperscript{2}, Abhay Tyagi\textsuperscript{3}, Sundeep Lakhtakia\textsuperscript{2}, Mohan Ramchandani\textsuperscript{2},\\
  \bf Rakesh Kalapala\textsuperscript{2}, D.~Nageshwar Reddy\textsuperscript{2}\\[6pt]
  \normalfont\textsuperscript{1}Department of Artificial Intelligence, Asian Institute of Gastroenterology, India\\
  \normalfont\textsuperscript{2}Department of Gastroenterology, Asian Institute of Gastroenterology, India\\
  \normalfont\textsuperscript{3}University of Vermont Medical Center, USA\\
  \normalfont\textsuperscript{*}Corresponding author: \texttt{mr.tyagi@gmail.com}
}

\begin{document}

\maketitle

\input{abstract}

\input{section1_introduction}
\input{section2_related_work}

\input{section3_dataset_pipeline}

\input{section4_evaluation_framework}

\input{section5_results}

\input{section6_discussion}

\begin{ack}
This work received no external funding. The authors declare no competing interests.
\end{ack}

\bibliographystyle{plainnat}
\bibliography{references}

\appendix
\input{appendix_A_taxonomy_and_counts}

\input{appendix_B_detailed_results}

\input{appendix_C_consensus_reannotation}

\input{appendix_D_extended_related_work}

\end{document}

%% file: abstract.tex
\begin{abstract}
Endoscopic diagnosis unfolds over native video, where a single
positive finding during traversal of the gastrointestinal tract sets
off a multi-minute chain of reasoning over location, morphology,
severity and intervention. Existing endoscopy multimodal benchmarks
test only static images or single-disease videos, which miss the
temporal and clinical reasoning that real-world reporting demands. We
introduce EndoSCOP-V, a video-grounded multi-turn evaluation framework
for structured endoscopy reporting and clinical reasoning. It contains
1{,}307 real-world endoscopy and colonoscopy videos (63.2~h; 97\%
1080p) averaging ${\sim}$174~s, the time-scale at which endoscopists
actually reason about a finding. Its 20{,}274 question--answer turns
are instantiated deterministically from clinician-filled structured
endoscopy-reporting EHRs, so no per-case annotation is synthetic or
LLM-rephrased. They come from 1{,}033 clinician-reviewed templates,
about 18 per disease, that span a clinically aligned 9-task $\times$
30-sub-task reporting taxonomy over 54 World Endoscopy Organization
(WEO) MST 3.0 diseases and examine each disease substantially more
deeply than prior endoscopy benchmarks. As in real case mix, 22\% of
procedures carry co-existing pathologies, up to six per case. To model
the hierarchical and conditional nature of clinical reporting, a
Markovian context-pruned oracle-forcing protocol evaluates the
dialogue turn by turn, injecting the gold answer before each
subsequent turn so that per-turn reasoning is isolated from cascading
dialogue errors. Clinician-marked key-frame anchors further separate
temporal-localization from frame-level comprehension failures. A panel
of 15 board-certified endoscopists sets a 78\% strict-accuracy
baseline on 1{,}500 questions. Across 14 open-weight LMMs, the best
27B-class model reaches 59.8\% strict accuracy, falling to 53.4\% once
format-trivial disease-rejection items are excluded. On a
between-subjects spot-check, the best of 3 frontier proprietary models
trails the clinician panel by 23~pp on strict NDR accuracy. Sub-task
analysis reveals consistent gaps in quantitative measurement,
multi-variable grading and intervention reasoning on long-horizon
video. The data are available at
\url{https://doi.org/10.7910/DVN/HRHPEX} and the code at
\url{https://github.com/abhi25t/EndoSCOP-V}.
\end{abstract}

%% file: section1_introduction.tex
\section{Introduction}
\label{sec:intro}

\begin{figure}[h]
\centering
\includegraphics[width=0.95\linewidth]{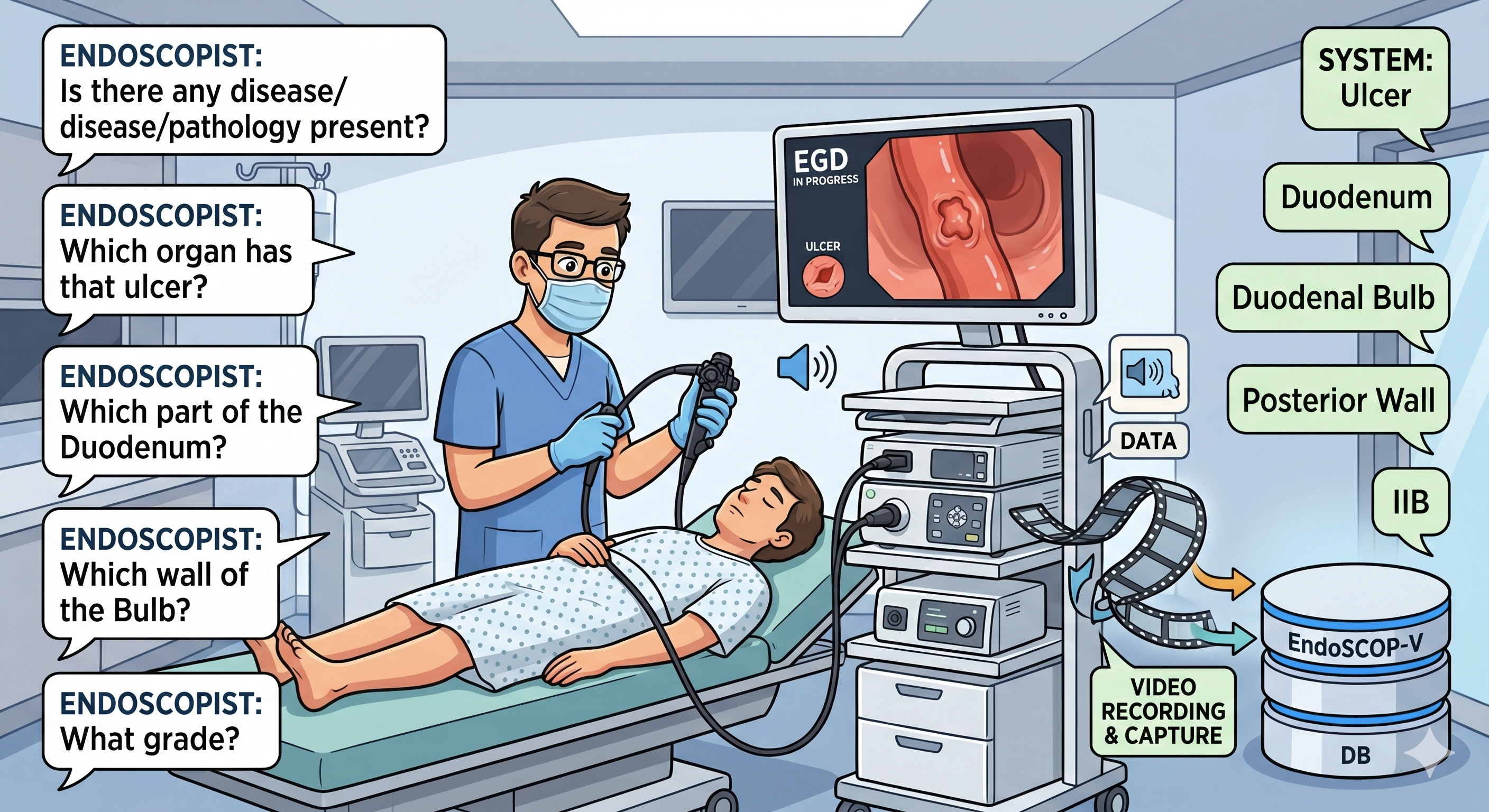}
\caption{Endoscopy reporting is inherently multi-turn. A single
finding cascades through a chain of follow-up questions (for example,
Disease? $\rightarrow$ Ulcer, Organ? $\rightarrow$ Duodenum,
Sub-location? $\rightarrow$ Duodenal Bulb, Wall? $\rightarrow$
Posterior, Grade? $\rightarrow$ IIB) that mirrors how an endoscopist
composes a structured EHR report. EndoSCOP-V evaluates models on this multi-turn dialogue
rather than on isolated single-shot recognition. Illustration
generated with Gemini 3.1 Pro.}
\label{fig:teaser}
\end{figure}

Gastrointestinal (GI) endoscopy, comprising esophagogastroduodenoscopy
(EGD) and colonoscopy, is the primary diagnostic modality for upper
and lower GI disease. A typical 8--15 minute procedure produces
several thousand high-resolution frames from which the endoscopist
localizes lesions, grades morphology and severity, decides on biopsy
or intervention, and composes a structured report that becomes the
patient's medical record. The clinical reasoning is inherently
\emph{multi-turn} (Figure~\ref{fig:teaser}), since a finding triggers
follow-up reasoning over location, morphology, grading and
intervention, where each answer depends on what was just seen. Recent
large multimodal models (LMMs) show competence on isolated endoscopy
frames, raising the question of whether they can support this
end-to-end reporting workflow at a clinically useful level.

The benchmarks that currently shape this question are poorly matched
to it. Public endoscopy datasets are predominantly image-level and
single-pathology, pairing each still frame with a one-shot label or a
single VQA turn. Where disease coverage is broader, the per-disease
examination is shallow, typically a single recognition decision per
image rather than the operator-level chain of follow-up questions on
location, morphology, grading and intervention that a clinical episode
requires. Ground-truth labels are often generated post hoc by an LLM
from captions, and evaluation protocols equate a correct answer with a
correct one-turn response. Such protocols have no notion of the
\textbf{conditional follow-up reasoning} that endoscopy depends on, in
which measurement, grading and intervention questions make sense only
once an earlier turn has established a finding.

We introduce \textbf{EndoSCOP-V (Endoscopic Screening and
Classification of Pathologies --- Video)}, an evaluation framework
and dataset for assessing LMMs across the full structured endoscopy
reporting workflow. EndoSCOP-V offers three contributions:

\begin{enumerate}[leftmargin=1.4em,itemsep=2pt,topsep=2pt]
\item \textbf{A Markovian, context-pruned oracle-forcing evaluation
protocol} that grades long multi-turn clinical dialogues turn by turn.
Because the gold answer to each earlier turn is forced into the
dialogue history, every downstream sub-task is graded on the model's
reasoning at \emph{that} turn, free of cascading errors from its own
earlier output. Conditioning on prior gold answers follows the
hierarchical radiology-reporting VQA
line~\cite{deperrois2025radvlm,pellegrini2023rad,pellegrini2025radialog}.
Our contribution is to combine this answer-reveal parent forcing with
Markovian per-turn context pruning on video-grounded endoscopy
dialogues, scored against targets instantiated deterministically from
clinician-filled EHR fields rather than mined from report text.
Conditional chains and disease-reveal prefixes further separate
follow-up detail extraction and intervention reasoning from disease
recognition (\S4).

\item \textbf{Per-disease reasoning depth.} An LLM-assisted converter drafts
questions from EHR field labels, and 15 board-certified endoscopists
review every stem and author every distractor, leaving 1{,}033
clinician-reviewed templates across a clinically aligned 9-task
$\times$ 30-sub-task reporting taxonomy (\S3). Across the 54 diseases
this yields \textbf{${\sim}$18 templates per disease on average},
covering detection, localisation, measurement, grading, conditional
follow-ups and intervention, substantially deeper than image-level
benchmarks. Per-case questions are then expanded
\emph{deterministically} from the templates and the doctor-filled case
report, which eliminates free-form LLM authoring and model-generated
ground truth.

\item \textbf{HD video grounding with per-disease key-frame anchors.}
EndoSCOP-V covers \textbf{54 diseases} drawn from the World Endoscopy
Organization (WEO) \emph{Minimal Standard Terminology
3.0}~\cite{aabakken2009minimal}, plus four non-disease findings. Its
\textbf{1{,}307 procedural video clips} (1{,}004 EGD + 303
colonoscopy) total \textbf{63.2 hours} at a mean length of
${\sim}$174~s, and \textbf{97\% are 1080p HD} at native acquisition
resolution. One or more clinician-marked key-frame anchors per disease
instance separate temporal-localization failures from comprehension
failures.
\end{enumerate}

Two further design choices set EndoSCOP-V apart from existing
benchmarks. First, \textbf{22\% of cases carry two or more co-existing
diseases} (216/58/15/1/1 cases with 2/3/4/5/6 diseases), mirroring the
case mix that endoscopists face and that single-pathology benchmarks
suppress. Second, the dataset ships with a \textbf{clinician baseline
of 78\% accuracy on a 1{,}500-question subset} and a
\textbf{five-rater consensus re-annotation of 100 cases (2{,}148
questions)}.

We evaluate 14 open-weight LMMs from the Qwen3~\cite{yang2025qwen3},
Hulu-Med~\cite{jiang2025hulu}, MedGemma~\cite{sellergren2025medgemma}
and ColonR1~\cite{ji2025colon} families on the 20{,}274-question
benchmark. The best open-weight model reaches \textbf{59.8\% strict
overall accuracy} and \textbf{53.4\%} once Disease-Rejection (DI-R)
items are excluded, both well below the 78\% clinician baseline. These
binary \emph{``is condition X present?''} probes for verified-absent
look-alikes prove format-trivial, since 9 of 14 open-weight models
exceed 90\% on them and the 3{,}213 items (15.8\% of all questions)
inflate aggregate scores. Clinicians show no such inflation (78\% on
both), so we report strict overall and DI-R-excluded (NDR) accuracy
throughout. In a between-subjects spot-check (\S3.3), the best of
three frontier proprietary models (Gemini~3-flash, ChatGPT 5.4-mini,
Claude Sonnet~4.6) trails the clinician panel by \textbf{23 percentage
points} on strict NDR accuracy. Sub-task analysis exposes systematic
weaknesses in quantitative measurement, multi-variable grading and
intervention recommendation that aggregate numbers hide.

%% file: section2_related_work.tex
\section{Related Work}
\label{sec:related}

EndoSCOP-V sits at the intersection of three lines of prior work that
have evolved in isolation, namely endoscopy multimodal benchmarks,
endoscopy report-generation and agent systems, and multi-turn or
EHR-grounded medical VQA. It contributes \textbf{per-disease reasoning
depth on HD video at clinical clip length}, instantiated
deterministically from a working endoscopy-reporting EHR.

Existing endoscopy benchmarks (PanEndoX~\cite{johri2025clinician},
EndoBench~\cite{liu2025endobench}, ColonVQA /
Colon-X~\cite{ji2025colon}, GI-Bench~\cite{zhu2026gi},
Kvasir-VQA-x1~\cite{gautam2025kvasir}) are image-only and single-turn,
and the closest \emph{video-grounded} peer,
\textbf{SurgViVQA}~\cite{drago2025surgvivqa}, is
\textbf{surgery-centric}, with only 2.88\% of its questions on polyp
diagnosis. Report-generation pipelines
(Report-Angel~\cite{jiang2026domain}, From Scope to
Script~\cite{kaklamanos2025scope}, Massimi~\cite{massimi2026large})
and agentic systems (EndoAgent /
EndoAgentBench~\cite{tang2025endoagent}) are primarily image-based and
evaluate on internal test sets. The closest methodological precedent
for our multi-turn protocol is the \textbf{hi-VQA / Rad-ReStruct /
RaDialog / RadVLM} radiology
line~\cite{deperrois2025radvlm,pellegrini2023rad,pellegrini2025radialog},
which models structured chest-X-ray reporting as hierarchical
autoregressive VQA with prior Q\&A as context. EndoSCOP-V extends its
3-level CXR hierarchy to a \textbf{$9\times30$ sub-task taxonomy on
multi-turn endoscopy video}, replaces autoregressive
prediction-propagation with gold-answer forcing, and grounds the
taxonomy in clinician-reviewed EHR templates.
Appendix~\ref{appx:related-work} surveys the broader landscape and
precedents.

\begin{table}[h]
\caption{Compact comparison of EndoSCOP-V with peer endoscopy
benchmarks. Appendix~\ref{appx:related-work}
(Table~\ref{tab:endo-bench-compare-full}) gives the full version with
an eval-items column.}
\label{tab:endo-bench-compare-compact}
\centering
\setlength{\tabcolsep}{3pt}
\small
\begin{tabular}{lllcc}
\toprule
Benchmark & Modality / Clip length & Diseases / Categories & Multi-turn & Multi-disease \\
\midrule
PanEndoX~\cite{johri2025clinician}                       & Image                          & 111 GI conditions             & N & N \\
EndoBench~\cite{liu2025endobench}                        & Image                          & 4 modalities                  & N & N \\
ColonVQA / Colon-X~\cite{ji2025colon}                    & Image                          & 76 findings (Colon only)      & N & N \\
GI-Bench~\cite{zhu2026gi}                                & Image                          & 20 lesions (EGD+Colon)    & N & N \\
SurgViVQA~\cite{drago2025surgvivqa}                      & 60 videos (8-frame)     & Polyps only; 20 Q-cats        & N & N \\
Kvasir-VQA-x1~\cite{gautam2025kvasir}                    & Image (perturbed)              & 5 categories (EGD+Colon)  & N & N \\
EndoAgentBench~\cite{tang2025endoagent}                  & Image (agentic)                & 5 diagnostic sub-tasks        & N & N \\
\textbf{EndoSCOP-V (ours)}                               & \textbf{HD video, mean ${\sim}$174~s} & \textbf{54 diseases}   & \textbf{Y} & \textbf{Y (22\%)} \\
\bottomrule
\end{tabular}
\end{table}

%% file: section3_dataset_pipeline.tex
\section{Dataset and Annotation Pipeline}
\label{sec:dataset}

We collected the data at the Asian Institute of Gastroenterology (AIG)
Hospital, Hyderabad, India, under Institutional Ethics Committee
approval AIG/IEC-Post BH\&R 81/03.2026-R1. Cases were sampled per
disease, targeting about 30 for each of the 54 diseases (Appendix
A.3), and because 35 of these occur only in the upper GI tract, the
corpus holds 1{,}004 EGD and 303 colonoscopy cases. Procedures were
recorded in native $1920\times1080$ HD on four Olympus endoscopes
(CV-170/190/290, CV-1500) and two pediatric scopes (XP170N, H190DL).
PHI overlays were cropped before frame extraction, yielding
$1350\times1080$ frames in the released corpus.

\subsection{Question taxonomy}
\label{sec:taxonomy}

Each EHR field corresponds to a question in one of nine task
categories, which divide into 30 sub-tasks (Table~\ref{tab:taxonomy},
with full descriptions in Appendix~A.1).

\begin{table}[h]
\caption{9-task $\times$ 30-sub-task reporting taxonomy.}
\label{tab:taxonomy}
\centering
\small
\begin{tabular}{lll}
\toprule
Task & Abbr. & Sub-tasks \\
\midrule
Disease Identification          & DI & DI-M, DI-C, DI-R, DI-N \\
Anatomical Localization         & AL & AL-O, AL-P, AL-W \\
Morphological Assessment        & MA & MA-T, MA-S, MA-C, MA-M, MA-SZ \\
Pathological Feature Assessment & PF & PF-D, PF-M, PF-U, PF-S, PF-B \\
Quantification                  & QN & QN-L, QN-S \\
Grading \& Scoring              & GS & GS-MC, GS-SS, GS-EC, GS-AG \\
Complication Recognition        & CF & CF-B, CF-S, CF-A \\
Clinical Reasoning              & CR & CR-E, CR-I \\
Action Recognition              & AR & AR-B, AR-V \\
\bottomrule
\end{tabular}
\end{table}

\textbf{Four sub-tasks are video-exclusive}, each defined by an event
that unfolds over time. \emph{Bleeding} (CF-B) needs temporal
information to tell oozing from spurting, \emph{Stricture \&
Obstruction} (CF-S) turns on traversability, a procedural event,
\emph{Biopsy} (AR-B) is an action recognized only in motion, and
\emph{Visualization} (AR-V) covers navigation events that span
seconds. The frame ablation (\S\ref{sec:results-ablation}) confirms
that the two action-recognition sub-tasks, AR-B and AR-V, are answered
better from video than from any single frame, even an oracle one, which makes EndoSCOP-V
video-native. The two complication sub-tasks, CF-B and CF-S, are, by
contrast, answered best from the clinician-marked keyframe. The other
26 sub-tasks can benefit from video without strictly requiring it, for
example through multi-angle parallax for size estimation (MA-SZ),
multi-frame light reflections for surface texture (MA-S),
$360^{\circ}$ sweeps for variceal column counts (QN-S) or retroflex
viewing for cardio-esophageal grades (GS-AG).

\textbf{Clinical reasoning in structured form.} EndoSCOP-V evaluates
clinical reasoning through the structured report rather than free
text. Structured reporting on a standard
terminology~\cite{aabakken2009minimal} is the recommended form of
endoscopic documentation~\cite{bretthauer2016requirements}, and its
fields encode the reasoning an endoscopist performs. The Clinical
Reasoning task asks for the suspected etiology or likely context
(CR-E) and for the final impression and disease activity (CR-I), which
integrate evidence across the procedure, while conditional chains
(\S3.2) add follow-up reasoning over location, morphology, grading and
intervention. Answers are verifiable choices, which keeps supervision
deterministic but leaves free-form narrative, calibrated uncertainty
and ranked differential diagnosis outside this scope (\S6).

\subsection{Annotation pipeline}
\label{sec:pipeline}

Templates are authored in two passes. First, an LLM-assisted converter
drafts candidate questions from the structured EHR field labels of
each entity, choosing the sub-task class, the MCQ option set and any
conditional dependencies on parent attributes. Fifteen board-certified
endoscopists then review every draft, deleting items that cannot be
recovered from procedural video alone (those requiring tactile
feedback or histopathology, for example), rewriting stems in clinical
idiom and authoring every distractor. The review keeps \textbf{1{,}033
of 1{,}588 drafted templates}, roughly 18 per disease, covering
detection, localization, grading, measurement, conditional follow-up
reasoning and intervention. Each of the 58 entities (54 diseases + 4
non-disease findings) maps to a single YAML template that anchors all
per-case question authoring.

Per-case ground truth comes from the same 15 endoscopists. Each filled
the hospital's structured EHR endoscopy report template
retrospectively for a disjoint subset of the 1{,}307 cases, using the
fields and fill-in logic of routine clinical practice, with access to
reference material (internet, textbooks, endoscopic atlases). Per-case
QAs are then produced \textbf{deterministically} from the
doctor-reviewed templates and the doctor-filled report. Templates are
matched to the report's recorded diseases and non-disease findings,
questions instantiate only when their required EHR fields are
populated, and conditional questions are emitted only for triggering
parent values, with the parent's answer prefixed onto the child
(oracle forcing, \S4). The resulting multi-turn dialogue follows the
report's reading sequence (insertion $\rightarrow$ withdrawal
$\rightarrow$ final assessment). This pipeline yields \textbf{20{,}274
multi-turn QAs across 1{,}307 cases} (mean 15.5 turns per case, median
14, range 1--73), fully reproducible from templates and case-report
files.

Some templates encode parent--child question chains, in which a child
question fires only when its parent attribute takes specific values.
\emph{``How large is the largest ulcer?''}, for instance, is
meaningful only after \emph{``are ulcers present?''} has been answered
``Yes''. The oracle-forcing protocol (\S4) discloses the parent's
answer at the start of the child's turn, so the child is graded purely
on what it tests rather than on whether the model remembers the
parent. Organ, sub-location and organ-wall chains (sub-tasks AL-P /
AL-W) fire in most disease cases, and Appendix~A.5 works through
disease-specific chains.

\subsection{Cohort properties}
\label{sec:cohort}

\paragraph{Multi-disease cases.} 22\% of cases (291/1{,}307) carry two or more strict diseases, and a
single procedure can hold up to six. In these cases the templates of
every present disease are concatenated, the dialogue follows the EHR's
reading sequence, and DI-R distractors are drawn at the case level, so
a clinically plausible look-alike for disease A cannot accidentally be
a correct answer for a co-occurring disease B.

\paragraph{Clinician baseline and evaluation sets.} Results are reported on three sets. \textbf{Set A} is the full
benchmark (20{,}274 questions), answered by every open-weight model
except ColonR1, which is scored on its 9{,}897-question
eligibility-filtered subset. \textbf{Set B} is the clinician baseline,
in which each of the 15 endoscopists answered, closed-book, 7 randomly
drawn cases excluding their own annotated cases (${\sim}$100 questions
each, ${\sim}$1{,}500 in total from 105 cases) and achieved
\textbf{77.9\% strict accuracy} against EHR ground truth. \textbf{Set
C} covers three frontier proprietary models, each answering its own
random draw of 7 cases from the full benchmark (${\sim}$100
questions). Because these cases differ from the clinicians', Set C
comparisons with the clinician baseline are between-subjects and
directional.

\subsection{Consensus re-annotation}
\label{sec:consensus}

To bound the label noise in single-rater ground truth, we re-annotated
a stratified subset of 100 cases with five independent raters per
question. They were the endoscopist who filled the case's EHR report
and four other endoscopists from the same panel who had not seen the
case, and each answered every question blind to the others and to the
recorded label. The subset combines 75 cases sampled to match the full
benchmark on procedure mix and task share with 25 cases over-sampled
for the lowest-agreement sub-tasks (grading, measurement, etiology). A
question receives a consensus label when at least three of the five
raters choose the same answer, which labels 2{,}044 of the subset's
2{,}148 questions and excludes the 104 ties.

Agreement on the subset is \textbf{Krippendorff's $\alpha$ = 0.77} and
\textbf{Fleiss' Kappa = 0.76}. We report $\alpha$ alongside Kappa
because it accommodates the items a rater left unanswered. All five
raters agree on 1{,}355 of the 2{,}044 labelled questions, four on 305
and three on 384. The recorded EHR label matches the consensus on
\textbf{90.7\%} of questions (1{,}854/2{,}044), within the
87.5--91.5\% range of the four other raters. Disagreement concentrates
on inherently subjective items (descriptive morphology, multi-variable
grading, borderline measurements, etiological inference). Even there,
most sub-tasks lie between 0.57 and 0.83 and the lowest is 0.44
(anatomical grading), while the lowest task, clinical reasoning,
reaches 0.59 (Appendix~\ref{appx:consensus}).

Scoring the models of \S5 against the consensus instead of the
recorded labels changes each model's accuracy by at most 1.5~pp (mean
$|\Delta|$ = 0.8~pp). The ranking holds (\textbf{Spearman $\rho$ =
0.96}, 73 of 78 model pairs concordant), with the top and bottom of
the table unchanged, so model accuracy and ranking are invariant to
the label source.

EndoSCOP-V is released under CC BY-NC-SA~4.0.

%% file: section4_evaluation_framework.tex
\section{Evaluation Framework}
\label{sec:eval}

Two design choices distinguish this protocol, namely Markovian
per-turn context pruning, which prevents cascading dialogue errors,
and oracle forcing of parent answers through \emph{answer-reveal
prefixes}, so that child questions test clinical reasoning. The full
evaluation pipeline is open-sourced.

\subsection{Model input and turn protocol}
\label{sec:eval-input}

Every model is fed uniformly sampled frames and never the
clinician-marked keyframe, which serves only in dataset construction
to locate the clip and anchor question generation. The frame budget is
set per tier by each model's context window. Models with large
contexts (the Qwen family at 256k tokens and MedGemma at 128k) receive
100 uniformly sampled frames. The medical specialist models Hulu-Med
(16k) and ColonR1 (32k) receive 16, matched to the tighter of the two
so that both are compared on identical input. Cross-tier comparisons
therefore confound frame count with model capability. All models share
a \emph{fixed} system prompt, ``an expert gastroenterologist and
endoscopist reviewing a \{procedure\_type\} video'', so that
capability is not confounded with prompt-tuning effort, and frame
numbers rendered in green in the top-left corner let the model
reference timestamps. Each turn is then built deterministically from
the QA file as an \emph{optional} disease- or answer-reveal prefix,
followed by the question stem, a letter-indexed MCQ set (or
``Yes\,/\,No'' for binary) and a JSON-schema reply instruction. The
model returns a JSON object matching the schema, with a regex fallback
on parse failure (\S\ref{sec:eval-format}).

Each dialogue runs in three phases. Case-level \textbf{Disease
Identification} comes first (DI-M, then DI-C\,/\,DI-R\,/\,DI-N
shuffled, no prefixes). Per-disease \textbf{Anatomical context} then
opens each disease block when AL is present, with the
\emph{disease-reveal prefix} on AL-O and AL-O revealed for
AL-P\,/\,AL-W, and if AL is absent the prefix shifts to the first turn
of Phase 3. Last comes per-disease \textbf{Detail}
(MA\,/\,PF\,/\,GS\,/\,CF\,/\,CR\,/\,AR in EHR reading order), whose
conditional children carry answer-reveal prefixes (\S3.2). This
isolates the \emph{recognition} signal (Phase 1, no prefixes) from the
\emph{reasoning} signal (Phases 2--3, disease forced).

\subsection{Markovian context-pruned grading}
\label{sec:eval-markov}

A model that answers turn $t$ incorrectly does not corrupt the input
at turn $t{+}1$, because the prefix of turn $t{+}1$ injects the
\emph{correct} parent answer. This is what we mean by \textbf{oracle
forcing}, in which each turn is graded as if every prior turn had been
answered correctly. We deliberately trade ``naturalistic
conversation'', where errors accumulate, for clean per-turn
measurement of clinical reasoning, because the benchmark aims to
characterize \emph{what models know} about endoscopy reporting rather
than \emph{how reliably they sustain a long dialogue under their own
errors}. Our reported accuracies therefore describe per-turn clinical
reasoning and should not be read as end-to-end performance on an
unassisted multi-turn dialogue, which is expected to be lower.

\subsection{Answer formats and scoring}
\label{sec:eval-format}

EndoSCOP-V uses three answer formats with deterministic scoring
rules:

\begin{itemize}[leftmargin=1.4em,itemsep=2pt,topsep=2pt]
\item \textbf{Binary} (Yes\,/\,No): exact match, 0 or 1.
\item \textbf{Single-MCQ} (one of 3 or 4 letters): exact match, 0 or
1.
\item \textbf{Multi-select MCQ} (a subset of a letter set, e.g., naming
which colonic segments contain ulcers): scored both \emph{strictly}
and \emph{partially}.
  \begin{itemize}[leftmargin=1.4em,itemsep=2pt,topsep=2pt]
  \item \textbf{Strict} $=$ 1 iff predicted set equals gold set
  exactly, else 0.
  \item \textbf{Partial} $= \max\!\bigl(0,\;\tfrac{|S_p \cap S_g|}{|S_g|} -
  \tfrac{|S_p \setminus S_g|}{2\,|S_g|}\bigr)$,
  i.e., $+1/n$ per correct selection ($n = |S_g|$), $-1/(2n)$ per
  spurious selection, floored at 0. This rewards partial recall while
  penalizing hallucinated extras at half rate.
  \end{itemize}
\end{itemize}

Answer extraction is strict JSON parsing, and parse failures fall back
to a regex that pulls \texttt{"answer": "\ldots"} and re-validates it
against the schema's enum. Predictions still unrecoverable after
fallback are scored 0. We report strict exact-match as the primary
metric throughout the paper, because a clinical reporting system
either records the correct answer or it does not. Partial credit runs
as a parallel column, since medical-VQA ground truth carries
irreducible inter-rater variability (\S\ref{sec:consensus}) and a
model's near-miss on a multi-select set is the kind of partial answer
two clinicians would often produce.

\subsection{Reporting}
\label{sec:eval-reporting}

Per-question scores are aggregated for each model at six
granularities. Procedure (endoscopy, colonoscopy, both), task (9),
sub-task (30) and disease levels each carry six metrics (binary,
single-MCQ, multi-MCQ strict, multi-MCQ partial, total strict, total
partial), and task and disease totals collapse the hierarchy for
compact between-model comparison.

\subsection{Frame-ablation protocol}
\label{sec:eval-ablation}

To test which sub-tasks require temporal input, we re-evaluated all
2{,}256 questions of the four video-exclusive sub-tasks
(\S\ref{sec:taxonomy}), on the 1{,}081 cases that carry at least one.
Each question was posed under four input conditions: a single
clinician-marked keyframe, 25 and 50 uniformly sampled frames, and the
full video. The ablation covers the five Qwen-family models with a
native video path, since MedGemma and Hulu-Med accept only image
lists. The keyframe condition is an oracle, because it relies on the
clinician's keyframe annotation, which no deployed system has and no
model receives in the main evaluation (\S\ref{sec:eval-input}).

%% file: section5_results.tex
\section{Results}
\label{sec:results}

We report strict exact-match accuracy (\S\ref{sec:eval-format}) on
Sets~A, B and C (\S\ref{sec:cohort}), with per-task results in
\S\ref{sec:results-task} and the full per-sub-task and per-disease
tables in Appendix~B.

\subsection{Aggregate and per-procedure leaderboard}
\label{sec:results-leaderboard}

\begin{table}[h]
\caption{Accuracy (\%). Best per column in \textbf{bold}.}
\label{tab:leaderboard}
\centering
\small
\setlength{\tabcolsep}{3pt}
\begin{tabular}{rlrrrrrr}
\toprule
Rank & Model & Strict Overall & Strict NDR & Partial Overall & Partial NDR & EGD & Colonoscopy \\
\midrule
1  & Qwen3.6-27B               & \textbf{59.8} & \textbf{53.4} & 62.6 & \textbf{57.0} & 60.5 & 57.2 \\
2  & Qwen3.5-9B                & 58.4 & 51.6 & 62.0 & 56.5 & 60.3 & 52.8 \\
3  & Qwen3.5-27B               & 58.2 & 52.2 & 61.2 & 56.0 & 58.3 & 56.2 \\
4  & Qwen3-VL-32B              & 56.4 & 48.8 & 60.3 & 53.3 & 56.5 & 53.9 \\
5  & Hulu-Med-32B              & 55.0 & 46.9 & 58.8 & 51.2 & 55.9 & 52.5 \\
6  & Hulu-Med-14B              & 54.6 & 47.0 & 56.9 & 49.2 & 55.5 & 51.4 \\
7  & Qwen3-VL-8B               & 54.2 & 46.0 & 58.4 & 51.1 & 54.9 & 51.5 \\
8  & Hulu-Med-7B               & 54.2 & 45.8 & 57.1 & 49.0 & 55.0 & 50.9 \\
9  & Hulu-Med-4B               & 51.7 & 43.2 & 56.0 & 48.3 & 52.5 & 49.0 \\
10 & MedGemma-27B              & 51.6 & 43.0 & 54.4 & 47.2 & 52.3 & 47.6 \\
11 & Qwen3.5-4B                & 51.5 & 46.9 & 55.2 & 50.9 & 53.8 & 45.1 \\
12 & MedGemma-4B               & 46.7 & 37.9 & 51.3 & 43.3 & 48.0 & 43.0 \\
13 & MedGemma-1.5-4B           & 44.5 & 38.2 & 48.0 & 42.2 & 43.9 & 45.1 \\
14 & ColonR1\textsuperscript{$\dagger$} & 43.4 & 35.6 & 46.5 & 39.2 & 46.9 & 40.1 \\
\midrule
\multicolumn{8}{l}{\emph{Set~C --- proprietary spot-check}} \\
   & Gemini~3-flash       & 61.4          & 55.1          & \textbf{63.3} & 57.4          & \textbf{62.3} & \textbf{59.9} \\
   & ChatGPT 5.4-mini     & 59.4          & 53.2          & 62.7          & 56.6          & 60.4          & 57.4 \\
   & Claude Sonnet~4.6    & 58.3          & 52.4          & 61.2          & 56.1          & 59.1          & 56.5 \\
\midrule
\multicolumn{8}{l}{\emph{Set~B --- clinician baseline}} \\
   & \textbf{Clinician panel} & \textbf{77.9} & \textbf{77.9} & \textbf{79.1} & \textbf{79.2} & \textbf{78.7} & \textbf{77.1} \\
\bottomrule
\end{tabular}
\\[2pt]
\begin{minipage}{\textwidth}
\footnotesize
$\dagger$~ColonR1 is scored on its eligibility-filtered subset
(9{,}897 questions across 668 cases from the 28 disease classes it was
trained for). 95\% CIs are under ${\pm}1$~pp on Set-A overall
accuracy, ${\pm}2.1$~pp on Set~B and ${\pm}{\sim}10$~pp on Set-C
cells.
\end{minipage}
\end{table}

\paragraph{Findings.}
\begin{itemize}[leftmargin=1.4em,itemsep=2pt,topsep=2pt]
\item \textbf{Qwen3.6-27B is the strongest open-weight model} at
59.8\% strict\,/\,53.4\% NDR overall. The 23~pp NDR gap to the
clinician panel (77.9\%) is the headline figure.
\item \textbf{The Qwen family dominates the top four} (Qwen3.6-27B,
Qwen3.5-9B, Qwen3.5-27B, Qwen3-VL-32B). Qwen3.5-9B (58.4\% strict)
even outperforms the larger Qwen3.5-27B (58.2\%) and Qwen3-VL-32B
(56.4\%), so scale alone does not predict ranking on EndoSCOP-V.
\item \textbf{Endoscopy training-data exposure dominates over generic
medical-tuning.} Neither the Hulu-Med nor the MedGemma family cracks
the top four, and the strongest medical-tuned model, Hulu-Med-32B at
55.0\% strict, trails Qwen3-VL-32B (56.4\%) at the same parameter
count. This reflects Hulu-Med's training on some endoscopy data,
whereas MedGemma's training mix is reported as not containing
endoscopy specifically.
\item \textbf{Every model except MedGemma-1.5-4B does better on EGD than on
colonoscopy}, by 2--9~pp, a direction consistent across 13 of 14
models, so EndoSCOP-V's colonoscopy split is genuinely harder for
current LMMs at the question level. The Set-C proprietary cohort shows
the same pattern (Gemini~3-flash 62.3 EGD\,/\,59.9 Col), as do the
Set-B clinicians (78.7\,/\,77.1).
\item \textbf{ColonR1, the only colonoscopy-specialised model, performs
\emph{worse} on colonoscopy (40.1\%) than on EGD (46.9\%) within its
eligibility subset} and ranks last overall at 43.4\% strict. It trails
the generalist Qwen3.6-27B on most diseases in the same subset. Read
with the generalist-beats-medical-tuned pattern above, narrow
colon-specialist fine-tuning on a 28-disease distribution sacrificed
more general-purpose visual reasoning than it gained in colon-specific
signal, which makes this a cautionary result for single-modality
medical-LMM specialisation.
\item \textbf{Frontier proprietary models trail the clinician panel by
${\approx}$23~pp on strict NDR} (Gemini~3-flash 55.1\% $\pm$
${\sim}$10~pp at $n$ ${\approx}$100, vs.~clinicians 77.9\%), and the
best frontier strict overall accuracy (61.4\%) exceeds the best
open-weight one (59.8\%) by only 1.6~pp.
\end{itemize}

\subsection{Per-task patterns}
\label{sec:results-task}

\begin{table}[h]
\caption{Strict accuracy (\%) by clinical task across the 14
open-weight models, where \emph{n} is the total number of questions
per task across the 1{,}307 cases.}
\label{tab:per-task}
\centering
\small
\setlength{\tabcolsep}{4pt}
\begin{tabular}{lrrrrrrrrr}
\toprule
Model & DI & AL & MA & PF & QN & GS & CF & CR & AR \\
\midrule
\emph{n}             & 6{,}585 & 1{,}390 & 3{,}387 & 3{,}503 & 609 & 1{,}176 & 1{,}701 & 1{,}026 & 897 \\
\midrule
Qwen3.6-27B          & 76.0          & \textbf{47.2} & \textbf{48.3} & 51.6          & \textbf{64.7} & 40.3          & 61.3          & 40.7          & 79.5 \\
Qwen3.5-9B           & 75.4          & 40.1          & 45.1          & \textbf{52.6} & 52.7          & 40.9          & 59.0          & 43.4          & 77.1 \\
Qwen3.5-27B          & \textbf{76.5} & 46.6          & 43.8          & 46.9          & 58.6          & 46.0          & 59.1          & 37.5          & \textbf{80.7} \\
Qwen3-VL-32B         & 74.9          & 41.8          & 42.5          & 44.2          & 58.0          & \textbf{46.5} & 57.4          & 35.1          & 75.5 \\
Hulu-Med-32B         & 66.8          & 39.6          & 45.0          & \textbf{52.6} & 54.4          & 42.9          & 60.5          & 38.7          & 72.8 \\
Hulu-Med-14B         & 70.4          & 29.5          & 47.7          & 49.0          & 52.5          & 38.5          & 58.9          & 35.1          & 70.2 \\
Qwen3-VL-8B          & 70.7          & 38.2          & 43.9          & 46.2          & 56.7          & 45.3          & 50.5          & 35.3          & 74.1 \\
Hulu-Med-7B          & 67.2          & 36.7          & 44.5          & 46.2          & 50.7          & 42.2          & \textbf{61.5} & \textbf{46.5} & 65.7 \\
Hulu-Med-4B          & 65.2          & 34.2          & 40.2          & 44.7          & 57.0          & 40.6          & 53.6          & 42.1          & 68.6 \\
MedGemma-27B         & 58.9          & 42.2          & 41.8          & 49.8          & 38.6          & 43.7          & 59.1          & 40.9          & 72.7 \\
Qwen3.5-4B           & 61.4          & 38.8          & 42.3          & 47.5          & 56.3          & 37.1          & 52.6          & 41.1          & 76.9 \\
MedGemma-4B          & 63.5          & 35.6          & 34.3          & 41.6          & 19.4          & 38.6          & 45.8          & 29.8          & 65.2 \\
MedGemma-1.5-4B      & 60.0          & 37.6          & 32.9          & 38.3          & 26.3          & 28.8          & 45.3          & 26.8          & 64.4 \\
ColonR1              & 61.7          & 27.1          & 36.6          & 35.7          & 29.4          & 23.6          & 44.6          & 23.7          & 64.6 \\
\bottomrule
\end{tabular}
\end{table}

\paragraph{Findings.}
\begin{itemize}[leftmargin=1.4em,itemsep=2pt,topsep=2pt]
\item \textbf{DI-M honest-recognition signal.} DI-M ($n{=}1{,}686$)
is the 4-option disease-identification MCQ, so chance is 25\%.
\textbf{The best open-weight model (Qwen3.5-27B, 58.9\%)} is 33.9~pp
above chance but well below the Set-B clinicians' overall DI accuracy
of 94.1\%. MedGemma-27B scores \textbf{24.8\% on DI-M, numerically
below chance}, which is what systematic answer-class bias looks like
on a 4-way recognition probe rather than a binary one.
\item \textbf{AR is the easiest task} (best 80.7\%, Qwen3.5-27B),
driven almost entirely by AR-B (biopsy-event recognition, with
sub-task accuracy of 80\%${+}$ for most models). AR-V (visualisation
events, $n{=}41$) is much weaker, subject to the small-$n$ caveat
below.
\item \textbf{Identification moderate, composition weak.}
Single-attribute recognition holds up (DI-M 58.9\%, AR-B 80\%${+}$),
but tasks composing multiple visual cues into a single categorical
output collapse uniformly across all 14 models. These include
segment-level wall position (AL best 47.2\%), multi-variable severity
bins such as Mayo / UCEIS / SES-CD / EREFS (GS 46.5\%) and
diagnostic-reasoning chains (CR 46.5\%). The same
\emph{recognise-but-fail-to-characterise} signature surfaces per
disease in Ulcerative Colitis (52.0\%) and Crohn's (36.6\%, best
open-weight), and this, not identification, is the deployment-relevant
capability gap.
\item \textbf{Best-per-task is distributed across families}, and no
single model dominates every task.
\item \textbf{Three sub-tasks have $n<50$} (DI-N${=}18$, QN-S${=}43$,
AR-V${=}41$). We report point estimates without confidence intervals
on these cells and make no model-vs-model claims at this granularity.
\end{itemize}

\subsection{Per-disease results}
\label{sec:results-disease}

Appendix~B (Table~\ref{tab:per-disease}) gives the full per-disease
strict-accuracy table for 58 entities (54 diseases + 4 non-disease
findings) and the eight selected open-weight models.

\paragraph{Findings.}
\begin{itemize}[leftmargin=1.4em,itemsep=2pt,topsep=2pt]
\item \textbf{21 of the 58 entities have every open-weight model
scoring under 58\%} (i.e., ${\geq}20$~pp below the 78\% Set-B
clinician baseline). The pattern concentrates on (a) inflammatory
bowel disease with multi-variable scoring (Crohn's, Ulcerative
Colitis), (b) subtle morphologic distinctions (Erosion, SRUS,
Submucosal Lesion), (c) tumor\,/\,mass identification (Tumor\,/\,Mass
best 48.6\%), and (d) most of the seven rare diseases.
\item \textbf{Per-disease ranking is stable across model families.}
The easiest diseases for the strongest model are Diverticulum
(80.0\%), Portal Hypertensive Gastropathy (76.0\%), Lymphangiectasia
(72.8\%), Polyp (72.1\%) and Xanthoma (71.4\%), all visually
distinctive, predominantly single-attribute findings. Easy diseases
for the strongest model are also easy for the second- and
third-ranked, while hard ones (Tumor\,/\,Mass, Extrinsic Compression,
IBD broadly) are uniformly hard across the leaderboard. This suggests
that the per-disease difficulty gradient reflects genuine
clinical-vision difficulty rather than idiosyncratic model blind
spots.
\end{itemize}

\subsection{Comparison with the clinician baseline}
\label{sec:results-clinician}

The clinician reference is Set B (\S\ref{sec:cohort}), which scores
\textbf{77.9\% strict accuracy} against EHR ground truth
($\pm$2.1~pp), overall and on NDR alike. Answering closed-book mirrors
routine reporting, where clinicians do not consult reference works
mid-procedure, and matches the models, which receive no retrieval,
whereas the ground truth was filled with reference material
(\S\ref{sec:pipeline}).

\paragraph{Reading the gap.} First, each endoscopist matches the five-rater consensus on only
87.5--91.5\% of questions (\S\ref{sec:consensus}), so the relevant
target is \textbf{not 100\% but the panel mean}. Subjective sub-tasks
(descriptive morphology, multi-variable grading, borderline
measurements, etiological inference), where rater agreement is lowest,
carry a tighter ceiling. Second, \textbf{the aggregate-vs-NDR gap is
itself a deployment signal}. Clinicians are flat across DI-R and the
rest (overall = NDR = 77.9\%), whereas every open-weight LMM shows a
${\sim}$5--8~pp gap. A model whose gap exceeds the panel's near-zero
gap will under-call genuine pathology when the differential includes a
clinically plausible look-alike.

\subsection{Frame ablation: keyframe, uniform frames and full video}
\label{sec:results-ablation}

\begin{table}[h]
\caption{Strict accuracy (\%) on the four video-exclusive
sub-tasks, mean over five models, equal weight per model. Paired
McNemar tests (video vs.\ keyframe, pooled over model $\times$
question outcomes) give $p < 0.001$ in all four sub-tasks.}
\label{tab:frame-ablation}
\centering
\small
\setlength{\tabcolsep}{4pt}
\begin{tabular}{lrrrrrr}
\toprule
Video-exclusive sub-task & \emph{n} & Keyframe (oracle) & 25 frames & 50 frames & Full video & Video $-$ keyframe \\
\midrule
AR-V & 41  & 40.5          & 50.7 & 51.2          & \textbf{54.3} & \textbf{+13.8} \\
AR-B & 856 & 72.5          & 78.6 & \textbf{80.2} & 79.1          & \textbf{+6.5} \\
CF-B & 718 & \textbf{82.0} & 77.9 & 77.2          & 76.2          & $-$5.8 \\
CF-S & 641 & \textbf{50.6} & 49.6 & 49.0          & 48.5          & $-$2.1 \\
\bottomrule
\end{tabular}
\end{table}

Table~\ref{tab:frame-ablation} shows two regimes. The two \emph{action-recognition}
sub-tasks, Biopsy (AR-B) and Visualization (AR-V), require temporal
input, since full video beats the keyframe by \textbf{+13.8~pp on
AR-V} and \textbf{+6.5~pp on AR-B}, with the same sign in all five
models. The gain does not require full video, because it is already
present at 25--50 uniformly sampled frames, which need no clinician
annotation, and AR-B peaks at 50~frames (80.2\%). The strongest model,
Qwen3.6-27B (rank 1 in Table~\ref{tab:leaderboard}), shows the largest
gains (AR-B +11.1, AR-V +26.2~pp), so the effect is not a small-model
artifact. The two \emph{complication} sub-tasks, Bleeding (CF-B) and
Stricture \& Obstruction (CF-S), run the other way. Both are best from
the single keyframe, again in all five models, yet 50 uniform frames
come within  ${\sim}$5~pp (CF-B) and  ${\sim}$1.6~pp (CF-S) of it. The
cost of not knowing where to look is therefore a few points rather
than task failure. Returns saturate early, as moving from one frame to
25 shifts accuracy by up to 10.2~pp, against at most 1.6~pp from 25 to
50.

Because the keyframe is an oracle, its numbers are an upper bound on
any deployable single-frame system, and even that bound loses both
action-recognition sub-tasks, so EndoSCOP-V contains sub-tasks that
need more than a single frame. The complication sub-tasks, by
contrast, are answered best from the clinician-marked keyframe, and
the benchmark separates items that need temporal reasoning from items
that do not.

%% file: section6_discussion.tex
\section{Limitations}
\label{sec:limitations}

The 23~pp NDR gap places current LMMs firmly in the \textbf{assistive}
rather than \textbf{replacement} regime for endoscopy reporting,
because a model that records 55\% of EHR ground-truth fields correctly
produces a draft that a reporting clinician must still verify and
correct, not a finalised report. \textbf{Single-institution,
single-rater per-case ground truth.} The 1{,}307 cases come from one
host hospital, and a single endoscopist filled the per-case EHR truth
for each case. A five-rater consensus on 100 cases agrees with the
recorded label on 90.7\% of questions and leaves the model ranking
unchanged (\S3.4), but the other 1{,}207 cases remain single-rater.
\textbf{Annotator--baseline overlap.} The 15 endoscopists who filled
per-case EHR ground truth (\S3.2) also provided the Set-B clinician
baseline (\S3.3), so the 78\% reference is a within-panel baseline
rather than an inter-panel benchmark. Because no endoscopist answered
a case they had annotated (\S3.3), recall of their own entries cannot
inflate the baseline, and residual familiarity is limited to the
benchmark's question phrasing. The consensus raters (\S3.4) come from
the same panel, so the consensus measures agreement within it rather
than across institutions. A panel drawn from a different institution
would be the definitive control for both and is the natural next step.
\textbf{Set~C is between-subjects, not paired.} Each proprietary model
answered its own ${\sim}$100 questions rather than the clinicians'
cases, so proprietary-vs-clinician comparisons are directional and
cannot support paired question-level significance tests.
\textbf{Pre-lavage, quick-movement clip selection.} EndoSCOP-V
deliberately samples pre-lavage frames, as in capsule endoscopy, where
lavage is physically impossible, so that scores remain comparable with
models trained on capsule corpora such as Kvasir-Capsule. It also
samples quick-movement segments to maximise mucosal coverage within
current LMM context budgets. Models evaluated on slow, lavaged,
well-lit clinical viewing would likely score higher. EndoSCOP-V's
number is therefore the more deployment-realistic estimate of what a
model would deliver in a live reporting workflow on an unfiltered
video stream, although it is \emph{not} a ceiling under all viewing
conditions. \textbf{Structured answers only.} Free-form narrative
generation, calibrated uncertainty expression and ranked differential
diagnosis are not evaluated, because deterministic, reproducible
supervision requires verifiable targets. These capabilities complement
the structured report (\S3.1) rather than replace it.

%% file: appendix_A_taxonomy_and_counts.tex
\section{Task Taxonomy and Corpus Counts}
\label{app:taxonomy}

This appendix gives the full task taxonomy
(Table~\ref{tab:taxonomy-full}, referenced from \S3.1), the
per-sub-task question counts (Table~\ref{tab:subtask-counts}, from \S3
and \S4.4), the per-disease coverage
(Table~\ref{tab:per-disease-coverage}, from \S3), the per-entity
template counts (Table~\ref{tab:per-entity-templates}, from \S3.2),
worked examples of the conditional-chain structure (\S\ref{app:chain-examples}, from \S3.2)
and a one-stop dataset summary (Table~\ref{tab:dataset-summary}, \S\ref{app:dataset-summary}). All counts are
computed directly from the released QA JSON files.

\subsection{Full task taxonomy}
\label{app:taxonomy-full}

EndoSCOP-V questions are organized into 9 task families and 30
sub-tasks. Each sub-task is named to match how endoscopy reporting
decomposes a finding: anatomical localization (where in the GI tract),
morphological assessment (what does the lesion look like),
pathological feature assessment (active disease features),
quantification (count and size), grading and scoring (clinical grading
scales such as Mayo, Forrest and Prague), complication recognition
(bleeding, stricture, associated findings), clinical reasoning
(etiology, impression and activity), and action recognition
(procedural events visible in the video). Disease identification is
treated as a separate task with four sub-tasks because it serves as
the gating Phase-1 task in evaluation (\S4.1).

Four sub-tasks are \emph{video-exclusive}, each defined by an event
that unfolds over time. The frame ablation
(\S\ref{sec:results-ablation}) confirms that the two
action-recognition sub-tasks are answered better from video than from
any single frame, even an oracle one, which makes EndoSCOP-V video-native rather than
image-native. The two complication sub-tasks, by contrast, are
answered best from the clinician-marked keyframe. The clinical
rationale for each:

\begin{itemize}
\item \textbf{CF-B (Bleeding).} Distinguishing oozing from spurting requires
temporal information.
\item \textbf{CF-S (Stricture \& Obstruction).} Traversability is a
procedural event observed only in motion.
\item \textbf{AR-B (Biopsy).} Biopsy is an action recognized only across
frames.
\item \textbf{AR-V (Visualization).} Anatomical-structure examination is a
navigation event spanning seconds.
\end{itemize}

The remaining 26 sub-tasks benefit from video without strictly
requiring it, for example through multi-angle parallax for size
estimation, multi-frame light reflections for surface texture,
$360^{\circ}$ sweeps for variceal column counts or retroflex viewing
for Hill's grade.

\begin{longtable}{@{}p{3.8cm}p{1cm}p{5cm}p{4.5cm}@{}}
\caption{Full task taxonomy (9 tasks $\times$ 30 sub-tasks). Bold
rows are video-exclusive sub-tasks.}
\label{tab:taxonomy-full} \\
\toprule
Sub-task & Abbr. & Description & Video advantage \\
\midrule
\endfirsthead
\multicolumn{4}{l}{\emph{(Table~\ref{tab:taxonomy-full} continued)}} \\
\toprule
Sub-task & Abbr. & Description & Video advantage \\
\midrule
\endhead
\midrule
\multicolumn{4}{r}{\emph{Continued on next page}} \\
\endfoot
\bottomrule
\endlastfoot
MCQ Identification        & DI-M  & Identify disease from 4 options                         & Multi-angle, temporal context \\
Confirmation              & DI-C  & Confirm suspected disease                               & Multi-angle, temporal context \\
Disease-Rejection         & DI-R  & Reject a clinically plausible look-alike not present in the case & Multi-angle, temporal context \\
Normal Assessment         & DI-N  & Identify absence of pathology at a location             & Full mucosal sweep of region \\
Organ                     & AL-O  & Upper-GI organ identification                           & Navigational context \\
Part / Segment            & AL-P  & Sub-organ segment localization                          & Scope entry into segment visible \\
Wall                      & AL-W  & Gastric / duodenal wall identification                  & Orientation cues from movement \\
Shape \& Type             & MA-T  & Sessile, flat, pedunculated, polypoid                   & Multi-angle reveals stalk \& base \\
Surface                   & MA-S  & Smooth, granular, nodular, ulcerated                    & Dynamic light reflections \\
Color                     & MA-C  & Lesion color assessment                                 & Multiple frames reduce lighting bias \\
Margins                   & MA-M  & Well-demarcated vs. ill-defined                         & Scope movement reveals border \\
Size                      & MA-SZ & Size estimation in mm bins                              & Depth cues from parallax \\
Distribution \& Extent    & PF-D  & Localized, patchy, diffuse, continuous                  & Scope traversal shows spread \\
Mucosal Features          & PF-M  & Erythematous, edematous, friable                        & Friability $=$ contact bleeding (temporal) \\
Ulcer / Erosion Features  & PF-U  & Aphthous, shallow, deep, undermined                     & Depth perception from multiple angles \\
Tissue Features           & PF-S  & Red signs, chronic changes, scarring                    & Red spots pulsate (temporal) \\
Contents \& Base          & PF-B  & Clean base, slough, fibrin, discharge                   & Fluid movement is temporal \\
Lesion Count              & QN-L  & Number of lesions along the tract                       & Sequential encounter during traversal \\
Structure Count           & QN-S  & Variceal columns, bands                                 & $360^{\circ}$ sweep reveals all columns \\
Morphological Classification & GS-MC & Paris, Kudo, NICE, JNET                            & Multi-angle for pit-pattern assessment \\
Severity Scoring          & GS-SS & Mayo, UCEIS, SES-CD, Forrest, Kodsi, Zargar, EREFS      & Bleeding component requires video \\
Extent Classification     & GS-EC & Prague, Montreal, LA Grade, Kimura--Takemoto            & Scope traversal measures extent \\
Anatomical Grading        & GS-AG & Hills, Ampulla Type, Goligher, Variceal Grade           & Retroflex / dynamic viewing \\
\textbf{Bleeding}         & \textbf{CF-B} & Active bleeding type, source                    & \textbf{Oozing vs. spurting is temporal --- exclusive} \\
\textbf{Stricture \& Obstruction} & \textbf{CF-S} & Presence, traversability, features      & \textbf{Scope passage is video-exclusive} \\
Associated Findings       & CF-A  & Surrounding mucosa, co-existing features                & Scope panning reveals surroundings \\
Etiology \& Context       & CR-E  & Suspected etiology or likely context                    & Integrates multiple temporal cues \\
Impression \& Activity    & CR-I  & Final clinical impression, disease activity             & Synthesizes full video evidence \\
\textbf{Biopsy}           & \textbf{AR-B} & Biopsy performed in video                       & \textbf{Procedural action --- exclusive} \\
\textbf{Visualization}    & \textbf{AR-V} & Anatomical structure examined in video          & \textbf{Navigation event --- exclusive} \\
\end{longtable}

\subsection{Per-sub-task question counts}
\label{app:subtask-counts}

Table~\ref{tab:subtask-counts} reports the number of questions per
sub-task across the full 1{,}307-case corpus, broken down by
procedure type (EGD vs. colonoscopy) and answer format (Binary /
Single-MCQ / Multi-MCQ). Three observations carry into the
evaluation framework (\S4.4):

\begin{itemize}
\item \textbf{DI-R is the largest sub-task} (3{,}213 questions, 15.85\% of
the dataset, 49\% of the DI task). Every case contributes a fixed-size
pool of Disease-Rejection binaries regardless of how many diseases are
present, so DI-R volume scales with case count rather than with
disease count. This motivates the DI-R-isolated reporting introduced
in \S1, because without isolation the format-trivial DI-R floor
inflates aggregate accuracy by 4.6--8.8~pp across open-weight models
(\S5.1).
\item \textbf{DI-N (18), AR-V (41), QN-S (43) are below 50 questions} and
are pooled to their parent task in headline reporting to keep cell
sizes interpretable, while the per-sub-task cells remain in Appendix~B
(Table~\ref{tab:per-subtask}) for completeness, with sample sizes
annotated.
\item \textbf{Some sub-tasks are dominated by one answer format.} AR-B is
92\% binary (was a biopsy taken, yes or no), QN-L is 53\% single-MCQ
size-bin and MA-T is 49\% multi-MCQ (selecting all applicable shape
descriptors). The \S4.3 partial-credit rule for multi-select is
therefore consequential in MA-T, MA-S, PF-M, PF-S, PF-B, CF-A, CR-E,
GS-MC and other multi-MCQ-heavy sub-tasks.
\end{itemize}

\begin{table}[h]
\caption{Per-sub-task question counts.}
\label{tab:subtask-counts}
\centering
\resizebox{\linewidth}{!}{%
\setlength{\tabcolsep}{4pt}
\footnotesize
\begin{tabular}{lrrrrrrrr}
\toprule
Sub-task & N & \% all & \% task & EGD & Colon & Binary & Single-MCQ & Multi-MCQ \\
\midrule
AL-O    & 348      & 1.72\%   & 25.04\%  & 348      & 0      & 46    & 297   & 5     \\
AL-P    & 713      & 3.52\%   & 51.29\%  & 679      & 34     & 49    & 503   & 161   \\
AL-W    & 329      & 1.62\%   & 23.67\%  & 302      & 27     & 32    & 235   & 62    \\
AR-B    & 856      & 4.22\%   & 95.43\%  & 720      & 136    & 788   & 68    & 0     \\
AR-V    & 41       & 0.20\%   & 4.57\%   & 18       & 23     & 41    & 0     & 0     \\
CF-A    & 342      & 1.69\%   & 20.11\%  & 256      & 86     & 121   & 16    & 205   \\
CF-B    & 718      & 3.54\%   & 42.21\%  & 401      & 317    & 244   & 320   & 154   \\
CF-S    & 641      & 3.16\%   & 37.68\%  & 498      & 143    & 263   & 287   & 91    \\
CR-E    & 558      & 2.75\%   & 54.39\%  & 438      & 120    & 31    & 262   & 265   \\
CR-I    & 468      & 2.31\%   & 45.61\%  & 347      & 121    & 144   & 261   & 63    \\
DI-C    & 1{,}668  & 8.23\%   & 25.33\%  & 1{,}340  & 328    & 1{,}668 & 0   & 0     \\
DI-M    & 1{,}686  & 8.32\%   & 25.60\%  & 1{,}353  & 333    & 0     & 1{,}686 & 0   \\
DI-N    & 18       & 0.09\%   & 0.27\%   & 13       & 5      & 18    & 0     & 0     \\
DI-R    & 3{,}213  & 15.85\%  & 48.79\%  & 2{,}555  & 658    & 3{,}213 & 0   & 0     \\
GS-AG   & 134      & 0.66\%   & 11.39\%  & 103      & 31     & 0     & 134   & 0     \\
GS-EC   & 144      & 0.71\%   & 12.24\%  & 116      & 28     & 0     & 85    & 59    \\
GS-MC   & 337      & 1.66\%   & 28.66\%  & 270      & 67     & 0     & 269   & 68    \\
GS-SS   & 561      & 2.77\%   & 47.70\%  & 369      & 192    & 80    & 481   & 0     \\
MA-C    & 427      & 2.11\%   & 12.61\%  & 223      & 204    & 63    & 255   & 109   \\
MA-M    & 301      & 1.48\%   & 8.89\%   & 211      & 90     & 159   & 74    & 68    \\
MA-S    & 667      & 3.29\%   & 19.69\%  & 537      & 130    & 88    & 288   & 291   \\
MA-SZ   & 682      & 3.36\%   & 20.14\%  & 494      & 188    & 41    & 641   & 0     \\
MA-T    & 1{,}310  & 6.46\%   & 38.68\%  & 1{,}129  & 181    & 387   & 283   & 640   \\
PF-B    & 528      & 2.60\%   & 15.07\%  & 447      & 81     & 205   & 121   & 202   \\
PF-D    & 523      & 2.58\%   & 14.93\%  & 299      & 224    & 46    & 277   & 200   \\
PF-M    & 1{,}223  & 6.03\%   & 34.91\%  & 987      & 236    & 435   & 190   & 598   \\
PF-S    & 830      & 4.09\%   & 23.69\%  & 551      & 279    & 312   & 253   & 265   \\
PF-U    & 399      & 1.97\%   & 11.39\%  & 192      & 207    & 198   & 42    & 159   \\
QN-L    & 566      & 2.79\%   & 92.94\%  & 423      & 143    & 265   & 301   & 0     \\
QN-S    & 43       & 0.21\%   & 7.06\%   & 43       & 0      & 3     & 40    & 0     \\
\midrule
\textbf{TOTAL} & \textbf{20{,}274} & 100.00\% & --- & \textbf{15{,}662} & \textbf{4{,}612} & \textbf{8{,}940} & \textbf{7{,}669} & \textbf{3{,}665} \\
\bottomrule
\end{tabular}}
\end{table}

\subsection{Per-disease coverage}
\label{app:per-disease-coverage}

Table~\ref{tab:per-disease-coverage} reports per-disease case counts
and question counts under \textbf{Convention A} (strict diseases for
the standalone-vs-multi split). A case is \emph{standalone} iff it
carries exactly one strict disease in \texttt{diseases\_found}.
Non-disease findings (Hills hiatus, Ampulla of Vater / Papilla,
Mucosa, Normal) tag along without affecting the standalone/multi
label, so their rows show standalone = 0 / multi = 0, and their cases
are accounted for under the strict diseases they co-occur with (or in
the ``non-disease finding only'' bucket of
Table~\ref{tab:dataset-summary}'s three-way breakdown). The
Disease-Rejection (DI-R) row aggregates all 3{,}213 case-level
rejection questions (\S3.3) into a single row so per-disease question
totals are not contaminated.

We targeted approximately 30 cases per disease, subject to the patient
volume available at the host hospital during the collection window. Of
the 54 strict diseases, 36 reach the  $\geq$ 30-case target and 47
reach  $\geq$ 20 cases (mean 29.9, range 5--52). Seven rare diseases
finished below 20 cases (Caustic Injury 18, Pyloric Stenosis 17,
Linitis Plastica 15, Pseudomembranous Colitis 13, Dieulafoy Lesion 7,
Duodenal Varices 7, M\'en\'etrier disease 5), all conditions for which
sufficient cases simply did not present during the window. We report
per-disease results separately and flag low-N cells throughout the
evaluation.

\begin{landscape}
\setlength{\tabcolsep}{3pt}
\setlength{\LTcapwidth}{\linewidth}
\footnotesize
\begin{longtable}{lrrrrrrrrrrr}
\caption{Per-disease coverage (Convention A). Sorted by total cases
descending. Rows for non-disease findings (Hills hiatus, Normal,
Ampulla, Mucosa) appear in italics, and their cases are also counted
under the strict diseases they co-occur with, which is why
per-disease cases sum to more than 1{,}307. The Disease-Rejection
(DI-R) row aggregates over all 1{,}297 cases that have $\geq 1$
DI-R question.}
\label{tab:per-disease-coverage} \\
\toprule
Disease & Cases & Standalone & Multi & EGD & Colon & Questions & Binary & Single-MCQ & Multi-MCQ & EGD Q & Colon Q \\
\midrule
\endfirsthead
\multicolumn{12}{l}{\emph{(Table~\ref{tab:per-disease-coverage} continued)}} \\
\toprule
Disease & Cases & Standalone & Multi & EGD & Colon & Questions & Binary & Single-MCQ & Multi-MCQ & EGD Q & Colon Q \\
\midrule
\endhead
\midrule
\multicolumn{12}{r}{\emph{Continued on next page}} \\
\endfoot
\bottomrule
\endlastfoot
Gastric Polyp                          & 52 & 20 & 32 & 52 & 0  & 669 & 203 & 383 & 83  & 669 & 0 \\
Reflux Esophagitis/GERD                & 52 & 25 & 27 & 52 & 0  & 401 & 98  & 202 & 101 & 401 & 0 \\
Duodenal Ulcer                         & 49 & 21 & 28 & 49 & 0  & 593 & 138 & 249 & 206 & 593 & 0 \\
Erosion                                & 47 & 13 & 34 & 47 & 0  & 350 & 204 & 77  & 69  & 350 & 0 \\
Diverticulum                           & 43 & 36 & 7  & 37 & 6  & 668 & 270 & 256 & 142 & 606 & 62 \\
Esophageal Varices                     & 42 & 0  & 42 & 42 & 0  & 371 & 176 & 167 & 28  & 371 & 0 \\
Hiatus Hernia                          & 42 & 0  & 42 & 42 & 0  & 165 & 83  & 82  & 0   & 165 & 0 \\
Candidiasis                            & 41 & 11 & 30 & 41 & 0  & 161 & 78  & 83  & 0   & 161 & 0 \\
Stricture                              & 41 & 19 & 22 & 33 & 8  & 473 & 130 & 251 & 92  & 352 & 121 \\
Tumor / Mass                           & 40 & 32 & 8  & 30 & 10 & 451 & 110 & 182 & 159 & 330 & 121 \\
Lymphangiectasia                       & 37 & 23 & 14 & 37 & 0  & 312 & 99  & 164 & 49  & 312 & 0 \\
Gastric Mucosal Atrophy                & 36 & 12 & 24 & 36 & 0  & 198 & 46  & 75  & 77  & 198 & 0 \\
Polyp                                  & 35 & 31 & 4  & 0  & 35 & 330 & 99  & 158 & 73  & 0   & 330 \\
Gastric Outlet Obstruction             & 34 & 16 & 18 & 34 & 0  & 461 & 232 & 99  & 130 & 461 & 0 \\
Gastropathy                            & 34 & 13 & 21 & 34 & 0  & 243 & 35  & 65  & 143 & 243 & 0 \\
Melanosis Coli                         & 34 & 30 & 4  & 0  & 34 & 233 & 34  & 199 & 0   & 0   & 233 \\
Gastric Varices                        & 33 & 13 & 20 & 33 & 0  & 219 & 62  & 81  & 76  & 219 & 0 \\
Subepithelial Lesions                  & 33 & 23 & 10 & 30 & 3  & 323 & 75  & 132 & 116 & 291 & 32 \\
Barrett's esophagus                    & 32 & 21 & 11 & 32 & 0  & 394 & 189 & 175 & 30  & 394 & 0 \\
Duodenal Bulb Deformity                & 32 & 8  & 24 & 32 & 0  & 349 & 198 & 59  & 92  & 349 & 0 \\
GAVE                                   & 32 & 13 & 19 & 32 & 0  & 155 & 63  & 61  & 31  & 155 & 0 \\
H. Pylori                              & 32 & 8  & 24 & 32 & 0  & 317 & 160 & 98  & 59  & 317 & 0 \\
Hemorrhoids                            & 32 & 28 & 4  & 0  & 32 & 346 & 34  & 158 & 154 & 0   & 346 \\
Inlet Patch                            & 32 & 29 & 3  & 32 & 0  & 319 & 127 & 160 & 32  & 319 & 0 \\
Parasites                              & 32 & 28 & 4  & 20 & 12 & 169 & 52  & 99  & 18  & 122 & 47 \\
SRUS                                   & 32 & 29 & 3  & 0  & 32 & 452 & 138 & 266 & 48  & 0   & 452 \\
Angioectasia                           & 31 & 18 & 13 & 25 & 6  & 404 & 158 & 223 & 23  & 343 & 61 \\
Achalasia                              & 31 & 29 & 2  & 31 & 0  & 577 & 238 & 308 & 31  & 577 & 0 \\
Gastric Ulcer                          & 31 & 10 & 21 & 31 & 0  & 355 & 84  & 150 & 121 & 355 & 0 \\
Portal hypertensive gastropathy        & 31 & 0  & 31 & 31 & 0  & 179 & 61  & 60  & 58  & 179 & 0 \\
Schatzki Ring                          & 31 & 16 & 15 & 31 & 0  & 229 & 84  & 110 & 35  & 229 & 0 \\
Ulcerative Colitis                     & 31 & 31 & 0  & 0  & 31 & 561 & 99  & 361 & 101 & 0   & 561 \\
Coeliac Disease                        & 30 & 30 & 0  & 30 & 0  & 169 & 109 & 30  & 30  & 169 & 0 \\
Esophageal Polyp                       & 30 & 27 & 3  & 30 & 0  & 332 & 149 & 148 & 35  & 332 & 0 \\
Polyposis Syndrome                     & 30 & 25 & 5  & 15 & 15 & 98  & 30  & 68  & 0   & 59  & 39 \\
Xanthoma                               & 30 & 27 & 3  & 30 & 0  & 237 & 89  & 147 & 1   & 237 & 0 \\
Duodenopathy                           & 29 & 11 & 18 & 29 & 0  & 182 & 30  & 78  & 74  & 182 & 0 \\
Esophageal Ulcer                       & 29 & 18 & 11 & 29 & 0  & 476 & 110 & 170 & 196 & 476 & 0 \\
Mucosal Nodule                         & 29 & 21 & 8  & 29 & 0  & 414 & 109 & 251 & 54  & 414 & 0 \\
Ulcer                                  & 29 & 18 & 11 & 0  & 29 & 470 & 76  & 228 & 166 & 0   & 470 \\
Duodenal Polyp                         & 28 & 17 & 11 & 28 & 0  & 264 & 55  & 153 & 56  & 264 & 0 \\
Eosinophilic Esophagitis               & 28 & 23 & 5  & 28 & 0  & 398 & 186 & 168 & 44  & 398 & 0 \\
Esophageal Web                         & 28 & 24 & 4  & 28 & 0  & 301 & 132 & 113 & 56  & 301 & 0 \\
Crohn's Disease                        & 27 & 23 & 4  & 6  & 21 & 459 & 165 & 218 & 76  & 50  & 409 \\
Extrinsic Compression                  & 25 & 19 & 6  & 18 & 7  & 231 & 59  & 93  & 79  & 177 & 54 \\
\emph{Hills hiatus}                    & 25 & 0  & 0  & 25 & 0  & 36  & 0   & 36  & 0   & 36  & 0 \\
Radiation Colitis/Proctitis            & 25 & 23 & 2  & 0  & 25 & 331 & 110 & 105 & 116 & 0   & 331 \\
Fistula                                & 23 & 19 & 4  & 16 & 7  & 267 & 101 & 96  & 70  & 193 & 74 \\
Caustic Injury                         & 18 & 12 & 6  & 18 & 0  & 174 & 77  & 78  & 19  & 174 & 0 \\
Pyloric Stenosis                       & 17 & 6  & 11 & 17 & 0  & 163 & 64  & 65  & 34  & 163 & 0 \\
\emph{Normal}                          & 16 & 0  & 0  & 11 & 5  & 36  & 18  & 18  & 0   & 26  & 10 \\
Linitis Plastica                       & 15 & 15 & 0  & 15 & 0  & 166 & 38  & 59  & 69  & 166 & 0 \\
Pseudomembranous Colitis               & 13 & 11 & 2  & 0  & 13 & 189 & 68  & 70  & 51  & 0   & 189 \\
\emph{Ampulla of Vater / Papilla}      & 7  & 0  & 0  & 7  & 0  & 89  & 64  & 12  & 13  & 89  & 0 \\
Dieulafoy Lesion                       & 7  & 7  & 0  & 5  & 2  & 51  & 7   & 37  & 7   & 39  & 12 \\
Duodenal Varices                       & 7  & 6  & 1  & 7  & 0  & 51  & 11  & 21  & 19  & 51  & 0 \\
M\'en\'etrier disease                  & 5  & 4  & 1  & 5  & 0  & 43  & 10  & 10  & 23  & 43  & 0 \\
\emph{Mucosa}                          & 5  & 0  & 0  & 5  & 0  & 7   & 3   & 4   & 0   & 7   & 0 \\
\midrule
\textbf{DI-R (case-level)} & \textbf{1{,}297} & 0 & 0 & \textbf{994} & \textbf{303} & \textbf{3{,}213} & \textbf{3{,}213} & 0 & 0 & \textbf{2{,}555} & \textbf{658} \\
\end{longtable}
\noindent{\footnotesize\emph{Italicized rows are the four non-disease findings, whose
`Standalone' and `Multi' columns are 0 by convention. A disease
occurring in two organs is counted twice. Per-disease question totals
plus DI-R = 17{,}061 + 3{,}213 = 20{,}274.}}
\end{landscape}

\subsection{Per-entity template counts}
\label{app:per-entity-templates}

The 1{,}033 clinician-reviewed templates (\S3.2) are distributed
across the 58 entities (54 strict diseases + 4 non-disease findings)
at a mean of \textbf{17.81 templates per entity} (range 1--49). The
format mix at the template level is \textbf{490 Binary (47.4\%) / 356
Single-MCQ (34.5\%) / 187 Multi-MCQ (18.1\%)}, which composes per-case
into the question-level format mix reported in
Table~\ref{tab:subtask-counts} (8{,}940 / 7{,}669 / 3{,}665 = 44.1\% /
37.8\% / 18.1\%). The binary share is lower at question level because
a single binary template (e.g., a DI-R rejection) instantiates one
question per case, while a multi-attribute template (e.g., MA-T shape)
may instantiate several.

\begin{longtable}{rlrrrr}
\caption{Per-entity template counts. Alphabetical by entity name.
Format columns are template counts, and the question-level format
mix appears in Table~\ref{tab:subtask-counts}. Italicized rows are the
four non-disease findings.}
\label{tab:per-entity-templates} \\
\toprule
\# & Entity & Binary & Single-MCQ & Multi-MCQ & Total templates \\
\midrule
\endfirsthead
\multicolumn{6}{l}{\emph{(Table~\ref{tab:per-entity-templates} continued)}} \\
\toprule
\# & Entity & Binary & Single-MCQ & Multi-MCQ & Total templates \\
\midrule
\endhead
\midrule
\multicolumn{6}{r}{\emph{Continued on next page}} \\
\endfoot
\bottomrule
\endlastfoot
1  & \emph{Ampulla of Vater / Papilla}        & 14 & 2  & 2  & 18 \\
2  & Angioectasia                             & 18 & 14 & 2  & 34 \\
3  & Barrett's esophagus                      & 9  & 6  & 1  & 16 \\
4  & Candidiasis                              & 4  & 3  & 0  & 7 \\
5  & Caustic Injury                           & 13 & 7  & 1  & 21 \\
6  & Coeliac Disease                          & 19 & 1  & 3  & 23 \\
7  & Crohn's Disease                          & 12 & 11 & 5  & 28 \\
8  & Dieulafoy Lesion                         & 6  & 9  & 1  & 16 \\
9  & Diverticulum                             & 34 & 11 & 4  & 49 \\
10 & Duodenal Bulb Deformity                  & 13 & 2  & 3  & 18 \\
11 & Duodenal Polyp                           & 4  & 6  & 2  & 12 \\
12 & Duodenal Ulcer                           & 5  & 6  & 5  & 16 \\
13 & Duodenal Varices                         & 11 & 4  & 3  & 18 \\
14 & Duodenopathy                             & 3  & 3  & 6  & 12 \\
15 & Eosinophilic Esophagitis                 & 14 & 6  & 2  & 22 \\
16 & Erosion                                  & 7  & 2  & 2  & 11 \\
17 & Achalasia                                & 10 & 11 & 1  & 22 \\
18 & Esophageal Polyp                         & 7  & 5  & 3  & 15 \\
19 & Esophageal Ulcer                         & 7  & 7  & 15 & 29 \\
20 & Esophageal Varices                       & 14 & 11 & 2  & 27 \\
21 & Esophageal Web                           & 18 & 4  & 3  & 25 \\
22 & Extrinsic Compression                    & 5  & 8  & 5  & 18 \\
23 & Fistula                                  & 25 & 9  & 4  & 38 \\
24 & Gastric Mucosal Atrophy                  & 4  & 2  & 5  & 11 \\
25 & Gastric Outlet Obstruction               & 17 & 3  & 4  & 24 \\
26 & Gastric Polyp                            & 6  & 12 & 3  & 21 \\
27 & Gastric Ulcer                            & 5  & 6  & 4  & 15 \\
28 & Gastric Varices                          & 12 & 4  & 3  & 19 \\
29 & Gastropathy                              & 3  & 2  & 7  & 12 \\
30 & GAVE                                     & 4  & 2  & 1  & 7 \\
31 & H. Pylori                                & 14 & 4  & 2  & 20 \\
32 & Hemorrhoids                              & 4  & 7  & 7  & 18 \\
33 & Hiatus Hernia                            & 4  & 2  & 0  & 6 \\
34 & \emph{Hills hiatus}                      & 0  & 1  & 0  & 1 \\
35 & Inlet Patch                              & 6  & 6  & 2  & 14 \\
36 & Linitis Plastica                         & 5  & 4  & 4  & 13 \\
37 & Lymphangiectasia                         & 5  & 5  & 1  & 11 \\
38 & Melanosis Coli                           & 3  & 6  & 0  & 9 \\
39 & M\'en\'etrier disease                    & 4  & 2  & 5  & 11 \\
40 & \emph{Mucosa}                            & 4  & 7  & 0  & 11 \\
41 & Mucosal Nodule                           & 6  & 13 & 2  & 21 \\
42 & \emph{Normal}                            & 2  & 1  & 0  & 3 \\
43 & Parasites                                & 4  & 5  & 1  & 10 \\
44 & Polyp                                    & 6  & 8  & 3  & 17 \\
45 & Polyposis Syndrome                       & 3  & 3  & 0  & 6 \\
46 & Portal hypertensive gastropathy          & 4  & 2  & 2  & 8 \\
47 & Pseudomembranous Colitis                 & 8  & 6  & 6  & 20 \\
48 & Pyloric Stenosis                         & 15 & 4  & 2  & 21 \\
49 & Radiation Colitis/Proctitis              & 8  & 5  & 9  & 22 \\
50 & Reflux Esophagitis/GERD                  & 4  & 4  & 3  & 11 \\
51 & Schatzki Ring                            & 9  & 4  & 1  & 14 \\
52 & SRUS                                     & 8  & 11 & 2  & 21 \\
53 & Stricture                                & 7  & 13 & 5  & 25 \\
54 & Subepithelial Lesions                    & 7  & 14 & 6  & 27 \\
55 & Tumor / Mass                             & 8  & 12 & 8  & 28 \\
56 & Ulcer                                    & 6  & 8  & 7  & 21 \\
57 & Ulcerative Colitis                       & 8  & 14 & 7  & 29 \\
58 & Xanthoma                                 & 5  & 6  & 0  & 11 \\
\midrule
   & \textbf{TOTAL}                           & \textbf{490} & \textbf{356} & \textbf{187} & \textbf{1{,}033} \\
\end{longtable}

The depth distribution is uneven by clinical scope.
\textbf{Diverticulum (49), Fistula (38) and Angioectasia (34)} carry
the deepest template menus, because their EHR reporting decomposes
into many distinct location, count, morphology and complication
sub-questions. \textbf{Hills hiatus (1) and Normal (3)} carry the
thinnest because they encode either a single classification grade
(Hills) or the absence of pathology (Normal). Multi-disease cases
concatenate multiple per-entity template menus, which produces the
long right tail of the per-case turn-count distribution (mean 15.5,
max 73, \S3.2).

\subsection{Worked conditional-chain examples}
\label{app:chain-examples}

Two real cases illustrate the two gate types described in \S3.2:

\paragraph{Yes/No gate, single-disease (case c1005, Ulcerative Colitis).} Parent question 10 (sub-task PF-U, Binary) asks \emph{``Are ulcers
present in this Ulcerative Colitis?''} and is answered Yes. Three
children fire under \texttt{Subsection(Ulcers Present = Yes)}, each
carrying the \texttt{answer\_reveal\_prefix} \emph{``Ulcers are
present.''} Of these, q11 asks for distribution across colonic
segments (Multi-MCQ, PF-U), q12 for the largest-ulcer size bin
(Single-MCQ, MA-SZ) and q13 for ulcer morphology (Multi-MCQ, PF-U).
This exact chain fires in 20 cases of the corpus.

\paragraph{Multi-class gate, single-disease (case e1533, Subepithelial Lesion).} Parent question 10 (CR-E, Multi-MCQ) asks \emph{``What would an
expert endoscopist suspect as the likely etiology of this
Subepithelial Lesion?''} and is answered Aberrant Pancreas (chosen
from {Aberrant Pancreas, GIST, Lipoma, Leiomyoma}). Child q11 fires
under \texttt{Section(Likely Etiology = Aberrant Pancreas)} with
prefix \emph{``The likely etiology is Aberrant Pancreas.''}, asking
for the Aberrant-Pancreas subtype (CR-E, Single-MCQ).

\subsection{Dataset summary}
\label{app:dataset-summary}

Table~\ref{tab:dataset-summary} reproduces a one-stop summary of
the released corpus. All entries are derivable from the per-disease
(\S\ref{app:per-disease-coverage}), per-sub-task
(\S\ref{app:subtask-counts}), and per-entity
(\S\ref{app:per-entity-templates}) counts above.

\begin{table}[h]
\caption{EndoSCOP-V dataset summary.}
\label{tab:dataset-summary}
\centering
\small
\begin{tabular}{lr}
\toprule
Quantity & Value \\
\midrule
Cases (videos)                                            & 1{,}307 \\
EGD / colonoscopy split                                   & 1{,}004 / 303 (76.8\% / 23.2\%) \\
Total video duration                                      & 63.2 hours \\
Mean clip length                                          & ${\sim}$174~s \\
1080p / 720p                                              & 1{,}269 / 38 (97.1\% / 2.9\%) \\
Diseases (+ non-disease findings)                         & 54 ($+$ 4) \\
Disease coverage $\geq 30$ cases / $\geq 20$ cases        & 36 / 47 (of 54) \\
Question templates                                        & 1{,}033 (from 1{,}588 drafted, 35\% attrition) \\
Reviewing endoscopists                                    & 15 \\
Total QA pairs                                            & 20{,}274 \\
Turns per case (mean / median / max)                      & 15.5 / 14 / 73 \\
Single-disease cases                                      & 992 (75.9\%) \\
Multi-disease cases                                       & 291 (22.3\%) \\
Non-disease-only cases                                    & 24 (1.8\%) \\
Multi-disease distribution (2/3/4/5/6)                    & 216 / 58 / 15 / 1 / 1 \\
Disease-Rejection (DI-R) questions                        & 3{,}213 (15.8\% of all QAs) \\
Consensus re-annotation                                   & \\
\quad Cases / questions                                    & 100 / 2{,}148 \\
\quad Raters per question                                  & 5 \\
\quad Consensus-labelled questions                         & 2{,}044 (104 ties excluded) \\
\quad Krippendorff's $\alpha$ / Fleiss' Kappa               & 0.77 / 0.76 \\
\quad Agreement of each rater with consensus               & 87.5--91.5\% \\
Clinician baseline (Set~B)                                & \\
\quad Endoscopists $\times$ cases each                      & 15 $\times$ 7 (own annotated cases excluded) \\
\quad Questions / unique cases                             & ${\sim}$1{,}500 / 105 \\
\quad Strict accuracy                                      & 77.9\% \\
\bottomrule
\end{tabular}
\end{table}

%% file: appendix_B_detailed_results.tex
\section{Detailed per-sub-task and per-disease results}
\label{app:detailed-results}

This appendix carries the detailed result tables that \S5 omits for
length, namely per-sub-task strict accuracy across all 30 EndoSCOP-V
sub-tasks and per-disease strict accuracy across all 58 entities (54
diseases + 4 non-disease findings). Both tables show the same eight
open-weight models, selected to keep the tables readable while
preserving the generalist-vs-specialist comparisons of \S5. They are
the top five by overall strict accuracy in
\S\ref{sec:results-leaderboard} (Qwen3.6-27B, Qwen3.5-9B, Qwen3.5-27B,
Qwen3-VL-32B, Hulu-Med-32B), the strongest and weakest medical-tuned
variants (MedGemma-27B and MedGemma-1.5-4B) and the
colonoscopy-specialised ColonR1.

\textbf{Per-model evaluation settings} (frame budgets, input modality,
decoding parameters) are documented in the open-sourced
\texttt{config.yaml}. \textbf{Output parseability} is uniformly high
across the open-weight set, with 13 of 14 models below 3\% unparseable
responses. MedGemma-1.5-4B is the outlier at 6.78\%, which contributes
to its rank-13 placement.

Tables~\ref{tab:per-subtask} / \ref{tab:per-disease} carry point estimates only, and rows where 95\% CIs
exceed $\pm$10~pp are flagged with $\dagger$. The seven rare diseases
(case $n<20$, namely Caustic Injury, Pyloric Stenosis, Linitis
Plastica, Pseudomembranous Colitis, Dieulafoy Lesion, Duodenal
Varices, M\'en\'etrier Disease) appear in Table~\ref{tab:per-disease}
for completeness. Their 95\% CIs exceed $\pm$15~pp, so we draw no
model-vs-model claims at this scale.

\begin{landscape}
\subsection{Per-sub-task strict accuracy}
\label{app:per-subtask}
\setlength{\tabcolsep}{4pt}
\setlength{\LTcapwidth}{\linewidth}
\footnotesize
\begin{longtable}{lrrrrrrrrr}
\caption{Per-sub-task strict accuracy (\%) on Set~A, where $n$ is the
number of questions per sub-task in the 20{,}274-question full-cohort
denominator. ColonR1 is evaluated on its 9{,}897-question
eligibility-filtered subset, so its sub-task-level denominators are
smaller than the cohort $n$ shown in the table.}
\label{tab:per-subtask} \\
\toprule
Sub-task & $n$ & Qwen3.6-27B & Qwen3.5-9B & Qwen3.5-27B & Qwen3-VL-32B & Hulu-Med-32B & MedGemma-27B & MedGemma-1.5-4B & ColonR1 \\
\midrule
\endfirsthead
\multicolumn{10}{l}{\emph{(Table~\ref{tab:per-subtask} continued)}} \\
\toprule
Sub-task & $n$ & Qwen3.6-27B & Qwen3.5-9B & Qwen3.5-27B & Qwen3-VL-32B & Hulu-Med-32B & MedGemma-27B & MedGemma-1.5-4B & ColonR1 \\
\midrule
\endhead
\midrule
\multicolumn{10}{r}{\emph{Continued on next page}} \\
\endfoot
\bottomrule
\endlastfoot
AL-O (Organ)                  & 348   & 54.9 & 45.4 & 52.0 & 47.4 & 42.5 & 37.4 & 31.9 & 35.6 \\
AL-P (Part / Segment)         & 713   & 50.2 & 44.0 & 46.4 & 47.4 & 37.3 & 42.2 & 41.4 & 25.9 \\
AL-W (Wall)                   & 329   & 31.9 & 25.5 & 39.5 & 21.3 & 39.8 & 43.5 & 33.7 & 20.5 \\
AR-B (Biopsy)                 & 856   & 82.2 & 81.6 & 84.0 & 80.5 & 74.9 & 72.0 & 62.2 & 66.7 \\
AR-V $\dagger$ (Visualization) & 41   & 48.8 & 43.9 & 43.9 & 34.1 & 36.6 & 65.9 & 68.3 & 34.8 \\
CF-A (Associated Findings)    & 342   & 39.5 & 42.7 & 44.7 & 44.4 & 45.6 & 46.2 & 25.7 & 22.5 \\
CF-B (Bleeding)               & 718   & 80.1 & 76.9 & 76.0 & 74.5 & 80.5 & 75.2 & 54.2 & 49.7 \\
CF-S (Stricture \& Obs.)      & 641   & 48.5 & 47.0 & 45.9 & 45.6 & 42.3 & 42.1 & 45.1 & 50.0 \\
CR-E (Etiology \& Context)    & 558   & 37.5 & 38.0 & 31.0 & 27.4 & 36.6 & 31.7 & 19.9 & 17.9 \\
CR-I (Impression \& Activity) & 468   & 44.7 & 50.0 & 44.9 & 44.9 & 41.0 & 52.6 & 35.3 & 33.8 \\
DI-C (Confirmation)           & 1{,}668 & 59.7 & 61.5 & 65.0 & 54.6 & 37.2 & 27.8 & 47.2 & 30.9 \\
DI-M (MCQ Identification)     & 1{,}686 & 55.4 & 54.4 & 58.9 & 51.1 & 33.6 & 24.8 & 37.8 & 35.2 \\
DI-N $\dagger$ (Normal Assessment) & 18 & 72.2 & 77.8 & 72.2 & 100.0 & 100.0 & 100.0 & 88.9 & --- \\
DI-R (Disease-Rejection)      & 3{,}213 & 92.9 & 92.0 & 89.2 & 95.2 & 97.9 & 93.1 & 77.0 & 90.7 \\
GS-AG (Anatomical Grading)    & 134   & 23.1 & 41.8 & 25.4 & 38.1 & 36.6 & 40.3 & 19.4 & 57.1 \\
GS-EC (Extent Classification) & 144   & 39.2 & 42.7 & 39.2 & 42.0 & 32.2 & 35.7 & 32.2 & 8.7 \\
GS-MC (Morph.\ Classification) & 337  & 56.1 & 43.9 & 57.0 & 51.6 & 44.2 & 46.6 & 27.3 & 18.6 \\
GS-SS (Severity Scoring)      & 561   & 36.8 & 40.0 & 49.6 & 47.3 & 47.5 & 46.1 & 32.9 & 24.6 \\
MA-C (Color)                  & 427   & 55.3 & 51.8 & 42.2 & 37.5 & 44.0 & 34.4 & 31.9 & 35.9 \\
MA-M (Margins)                & 301   & 63.1 & 69.8 & 65.1 & 64.8 & 67.4 & 62.5 & 55.1 & 56.2 \\
MA-S (Surface)                & 667   & 47.5 & 40.2 & 42.1 & 41.4 & 45.3 & 36.3 & 32.5 & 26.0 \\
MA-SZ (Size)                  & 682   & 48.1 & 46.0 & 47.8 & 42.4 & 34.0 & 38.7 & 34.9 & 36.4 \\
MA-T (Shape \& Type)          & 1{,}310 & 41.7 & 39.6 & 36.9 & 38.6 & 45.1 & 42.8 & 26.8 & 34.3 \\
PF-B (Contents \& Base)       & 528   & 48.3 & 56.1 & 48.9 & 33.1 & 58.9 & 53.8 & 34.1 & 46.7 \\
PF-D (Distribution \& Extent) & 523   & 50.2 & 39.8 & 45.8 & 33.3 & 41.0 & 42.7 & 28.0 & 31.3 \\
PF-M (Mucosal Features)       & 1{,}223 & 57.3 & 59.8 & 50.2 & 52.9 & 57.2 & 53.9 & 40.0 & 39.8 \\
PF-S (Tissue Features)        & 830   & 46.4 & 45.4 & 36.5 & 38.8 & 49.5 & 43.9 & 37.2 & 29.6 \\
PF-U (Ulcer Features)         & 399   & 52.3 & 56.5 & 57.5 & 56.8 & 48.5 & 51.8 & 51.3 & 28.6 \\
QN-L (Lesion Count)           & 566   & 68.2 & 55.3 & 61.3 & 61.0 & 58.7 & 41.0 & 27.4 & 29.4 \\
QN-S $\dagger$ (Structure Count) & 43 & 20.9 & 20.9 & 25.6 & 20.9 & 9.3  & 9.3  & 16.3 & --- \\
\end{longtable}
\noindent{\footnotesize $\dagger$~Sub-tasks with $n < 50$ (AR-V ${=}$ 41, DI-N
${=}$ 18, QN-S ${=}$ 43), whose 95\% CIs exceed $\pm$10~pp.}
\end{landscape}

\begin{landscape}
\subsection{Per-disease strict accuracy}
\label{app:per-disease}
\setlength{\tabcolsep}{4pt}
\setlength{\LTcapwidth}{\linewidth}
\footnotesize
\begin{longtable}{lrrrrrrrrr}
\caption{Per-disease strict accuracy (\%) on Set~A, where $n$ is the
number of questions per disease. Italicized rows are the four non-disease
findings (Ampulla of Vater / Papilla, Hills hiatus, Mucosa, Normal).
ColonR1 cells show ``---'' for the 32 of 58 entities outside its
eligibility subset. A question in a multi-disease case is attributed
to each disease present, so per-disease $n$'s may sum above the
20{,}274 per-question total. A disease occurring in two organs is
counted twice. DI-R items are routed to a separate case-level row,
so per-disease rows exclude DI-R scores.}
\label{tab:per-disease} \\
\toprule
Disease & $n$ & Qwen3.6-27B & Qwen3.5-9B & Qwen3.5-27B & Qwen3-VL-32B & Hulu-Med-32B & MedGemma-27B & MedGemma-1.5-4B & ColonR1 \\
\midrule
\endfirsthead
\multicolumn{10}{l}{\emph{(Table~\ref{tab:per-disease} continued)}} \\
\toprule
Disease & $n$ & Qwen3.6-27B & Qwen3.5-9B & Qwen3.5-27B & Qwen3-VL-32B & Hulu-Med-32B & MedGemma-27B & MedGemma-1.5-4B & ColonR1 \\
\midrule
\endhead
\midrule
\multicolumn{10}{r}{\emph{Continued on next page}} \\
\endfoot
\bottomrule
\endlastfoot
Angioectasia                                 & 405 & 60.2 & 60.5 & 65.4 & 51.1 & 43.2 & 52.3 & 33.8 & 40.8 \\
Barrett's esophagus                          & 398 & 68.8 & 49.7 & 54.0 & 50.5 & 47.2 & 45.0 & 43.0 & --- \\
Candidiasis                                  & 161 & 60.9 & 47.2 & 55.3 & 57.8 & 45.3 & 34.2 & 31.1 & --- \\
Caustic Injury $\dagger$                     & 174 & 42.0 & 54.6 & 43.1 & 53.4 & 46.0 & 39.7 & 37.4 & --- \\
Coeliac Disease                              & 172 & 44.2 & 50.0 & 29.7 & 20.9 & 51.7 & 30.2 & 16.9 & --- \\
Crohn's Disease                              & 459 & 36.6 & 44.0 & 40.5 & 44.0 & 42.0 & 46.4 & 36.6 & 31.9 \\
Dieulafoy Lesion $\dagger$                   & 53  & 35.8 & 41.5 & 43.4 & 43.4 & 37.7 & 34.0 & 43.4 & 32.1 \\
Diverticulum                                 & 669 & 80.0 & 70.6 & 73.2 & 65.8 & 60.5 & 58.9 & 49.5 & 49.9 \\
Duodenal Bulb Deformity                      & 349 & 45.6 & 53.6 & 46.4 & 51.9 & 50.4 & 51.9 & 46.4 & --- \\
Duodenal Polyp                               & 264 & 47.0 & 43.6 & 58.0 & 56.8 & 39.0 & 44.3 & 29.2 & 42.4 \\
Duodenal Ulcer                               & 593 & 40.1 & 42.0 & 41.8 & 27.8 & 44.2 & 33.6 & 15.0 & 25.5 \\
Duodenal Varices $\dagger$                   & 52  & 42.3 & 44.2 & 36.5 & 36.5 & 26.9 & 26.9 & 40.4 & --- \\
Duodenopathy                                 & 182 & 47.3 & 45.6 & 44.0 & 37.4 & 52.2 & 36.3 & 23.1 & --- \\
Eosinophilic Esophagitis                     & 398 & 59.8 & 73.4 & 70.6 & 69.1 & 54.3 & 66.1 & 61.6 & --- \\
Erosion                                      & 350 & 50.3 & 48.3 & 54.3 & 52.9 & 49.7 & 24.3 & 36.0 & 20.9 \\
Esophageal Motility Disorder/Achalasia       & 591 & 51.9 & 51.9 & 53.5 & 41.5 & 43.1 & 42.3 & 49.1 & --- \\
Esophageal Polyp                             & 338 & 64.8 & 64.5 & 58.0 & 63.9 & 53.6 & 47.6 & 47.6 & 49.1 \\
Esophageal Ulcer                             & 477 & 40.3 & 53.9 & 43.8 & 47.2 & 49.7 & 38.2 & 30.6 & 31.4 \\
Esophageal Varices                           & 371 & 52.3 & 61.2 & 45.0 & 51.5 & 45.0 & 47.4 & 29.4 & --- \\
Esophageal Web                               & 301 & 45.5 & 44.2 & 41.9 & 42.9 & 34.6 & 46.5 & 40.5 & --- \\
Extrinsic Compression                        & 231 & 25.5 & 32.5 & 35.1 & 23.4 & 29.4 & 28.6 & 26.8 & 24.5 \\
Fistula                                      & 267 & 64.0 & 54.7 & 60.7 & 51.7 & 42.7 & 53.2 & 35.6 & 39.3 \\
GAVE                                         & 162 & 41.4 & 64.2 & 56.2 & 46.3 & 40.7 & 27.2 & 27.8 & --- \\
Gastric Mucosal Atrophy                      & 198 & 56.1 & 41.9 & 54.0 & 47.5 & 47.0 & 33.3 & 32.3 & --- \\
Gastric Outlet Obstruction                   & 461 & 57.0 & 58.1 & 59.7 & 70.9 & 39.9 & 47.9 & 28.4 & --- \\
Gastric Polyp                                & 669 & 63.7 & 56.1 & 61.7 & 54.3 & 51.9 & 51.1 & 44.8 & 54.4 \\
Gastric Ulcer                                & 355 & 54.6 & 53.0 & 49.9 & 44.5 & 45.4 & 34.1 & 36.1 & 20.7 \\
Gastric Varices                              & 220 & 53.6 & 51.8 & 48.6 & 47.3 & 39.1 & 42.7 & 24.1 & --- \\
Gastropathy                                  & 243 & 37.4 & 42.8 & 40.7 & 41.6 & 52.7 & 43.6 & 22.6 & --- \\
H.\ Pylori                                   & 317 & 49.5 & 45.1 & 35.0 & 53.0 & 43.8 & 30.6 & 21.8 & --- \\
Hemorrhoids                                  & 346 & 51.2 & 56.6 & 46.8 & 41.6 & 53.5 & 47.1 & 52.0 & 36.4 \\
Hiatus Hernia                                & 165 & 64.8 & 87.3 & 68.5 & 84.2 & 72.1 & 78.2 & 75.2 & --- \\
Inlet Patch                                  & 319 & 43.9 & 50.8 & 49.2 & 53.0 & 64.3 & 62.4 & 53.3 & --- \\
Linitis Plastica $\dagger$                   & 166 & 47.6 & 42.8 & 39.2 & 24.7 & 54.2 & 44.6 & 43.4 & --- \\
Lymphangiectasia                             & 312 & 72.8 & 68.9 & 67.3 & 58.7 & 72.4 & 49.7 & 54.2 & 34.3 \\
Melanosis Coli                               & 259 & 56.0 & 62.9 & 56.0 & 45.6 & 58.7 & 30.1 & 34.0 & 32.8 \\
M\'en\'etrier Disease $\dagger$              & 43  & 37.2 & 20.9 & 37.2 & 25.6 & 32.6 & 14.0 & 34.9 & --- \\
Mucosal Nodule                               & 417 & 64.3 & 58.0 & 66.2 & 55.2 & 50.4 & 54.9 & 46.3 & 51.4 \\
Parasites                                    & 170 & 66.5 & 67.1 & 58.2 & 55.3 & 33.5 & 28.8 & 35.9 & 40.9 \\
Polyp                                        & 340 & 72.1 & 56.2 & 69.7 & 66.8 & 44.7 & 45.3 & 47.4 & --- \\
Polyposis Syndrome                           & 98  & 65.3 & 70.4 & 60.2 & 68.4 & 4.1  & 16.3 & 23.5 & 77.5 \\
Portal Hypertensive Gastropathy              & 179 & 76.0 & 65.4 & 74.9 & 49.2 & 63.1 & 51.4 & 17.3 & --- \\
Pseudomembranous Colitis $\dagger$           & 189 & 56.6 & 45.0 & 54.5 & 50.8 & 56.1 & 38.6 & 42.9 & 25.4 \\
Pyloric Stenosis $\dagger$                   & 163 & 54.6 & 62.6 & 45.4 & 58.9 & 36.8 & 43.6 & 47.9 & --- \\
Radiation Colitis / Proctitis                & 331 & 40.2 & 44.4 & 43.5 & 38.7 & 38.7 & 33.8 & 43.5 & 27.7 \\
Reflux Esophagitis / GERD                    & 394 & 68.0 & 70.3 & 69.3 & 66.0 & 51.8 & 39.3 & 42.1 & --- \\
SRUS                                         & 452 & 46.0 & 48.2 & 43.8 & 46.0 & 44.7 & 46.0 & 36.9 & 29.8 \\
Schatzki Ring                                & 233 & 60.3 & 61.1 & 46.6 & 48.3 & 40.6 & 42.7 & 56.4 & --- \\
Stricture                                    & 473 & 61.3 & 45.2 & 60.5 & 59.4 & 46.9 & 42.7 & 32.8 & 41.9 \\
Submucosal / Subepithelial Lesions           & 324 & 55.9 & 46.0 & 47.5 & 35.8 & 42.3 & 42.6 & 30.9 & 43.5 \\
Tumor / Mass                                 & 451 & 44.6 & 44.6 & 47.7 & 30.4 & 20.6 & 31.0 & 22.6 & 18.3 \\
Ulcer                                        & 486 & 48.4 & 37.4 & 44.2 & 46.7 & 52.3 & 38.3 & 39.5 & --- \\
Ulcerative Colitis                           & 565 & 52.0 & 37.0 & 49.7 & 43.4 & 45.1 & 35.9 & 38.4 & 19.5 \\
Xanthoma                                     & 238 & 71.4 & 55.0 & 69.3 & 55.0 & 47.5 & 53.4 & 46.6 & --- \\
\emph{Ampulla of Vater / Papilla}            & 89  & 71.9 & 56.2 & 49.4 & 60.7 & 69.7 & 79.8 & 51.7 & --- \\
\emph{Hills hiatus} $\dagger$                & 36  & 0.0  & 52.8 & 27.8 & 11.1 & 11.1 & 33.3 & 33.3 & --- \\
\emph{Mucosa} $\dagger$                      & 7   & 14.3 & 42.9 & 14.3 & 28.6 & 28.6 & 28.6 & 28.6 & --- \\
\emph{Normal}                                & 72  & 86.1 & 76.4 & 86.1 & 100.0 & 100.0 & 100.0 & 86.1 & --- \\
\end{longtable}
\noindent{\footnotesize $\dagger$~Rows where 95\% CIs exceed $\pm$10~pp. These fall into two
groups: (i) the seven rare diseases identified in \S A.3 (Caustic
Injury, Pyloric Stenosis, Linitis Plastica and Pseudomembranous
Colitis at 13--18 cases despite 163--189 questions, plus Dieulafoy
Lesion, Duodenal Varices and M\'en\'etrier Disease at 5--7 cases and
43--53 questions) and (ii) two non-disease findings with very small
question counts (\emph{Hills hiatus} $n$ = 36, \emph{Mucosa} $n$ = 7).
Per-question accuracy on these rows can swing 10--20~pp under random
subsampling and should not be read as model-comparable.}
\end{landscape}

%% file: appendix_C_consensus_reannotation.tex
\section{Consensus re-annotation: detailed results}
\label{appx:consensus}

This appendix gives the detail behind \S\ref{sec:consensus}.
Tables~\ref{tab:consensus-task} and~\ref{tab:consensus-subtask} report
inter-rater reliability per task and per sub-task on the 100-case
consensus subset (2{,}148 questions, five raters).
Table~\ref{tab:consensus-model} reports model accuracy against the
recorded EHR label and against the consensus label.

\subsection{Inter-rater reliability}
\label{appx:consensus-irr}

\begin{table}[h]
\caption{Reliability per task (five raters, 2{,}148 questions).
\emph{Questions} counts all questions of the task; \emph{Labelled}
counts those with a $\geq$3/5 consensus. Krippendorff's $\alpha$
treats unanswered items as missing; raw = mean pairwise agreement.}
\label{tab:consensus-task}
\centering
\small
\begin{tabular}{lrrrrr}
\toprule
Task & Questions & Labelled & $\alpha$ & Fleiss' Kappa & Raw \\
\midrule
AR Action Recognition & 92 & 92 & 0.91 & 0.91 & 0.95 \\
DI Disease Identification & 629 & 629 & 0.88 & 0.88 & 0.92 \\
QN Quantification & 76 & 76 & 0.78 & 0.78 & 0.82 \\
CF Complication Recognition & 169 & 159 & 0.78 & 0.78 & 0.81 \\
PF Pathological Feature Assessment & 383 & 354 & 0.71 & 0.71 & 0.75 \\
MA Morphological Assessment & 397 & 371 & 0.67 & 0.67 & 0.71 \\
AL Anatomical Localization & 140 & 132 & 0.67 & 0.66 & 0.73 \\
GS Grading \& Scoring & 150 & 135 & 0.63 & 0.63 & 0.73 \\
CR Clinical Reasoning & 112 & 96 & 0.59 & 0.59 & 0.65 \\
\midrule
\textbf{All} & \textbf{2{,}148} & \textbf{2{,}044} & \textbf{0.77} & \textbf{0.76} & \textbf{0.80} \\
\bottomrule
\end{tabular}
\end{table}

\begin{table}[h]
\caption{Reliability per sub-task (five raters; sub-tasks with at least 10
questions in the subset). AR-V (8 questions) is pooled into AR in
Table~\ref{tab:consensus-task}; DI-N does not occur in the subset.}
\label{tab:consensus-subtask}
\centering
\small
\begin{tabular}{lrrrr}
\toprule
Sub-task & Questions & $\alpha$ & Fleiss' Kappa & Raw \\
\midrule
DI-M MCQ Identification & 166 & 0.83 & 0.83 & 0.88 \\
DI-C Confirmation\textsuperscript{$\dagger$} & 166 & 0.20 & 0.20 & 0.88 \\
DI-R Disease-Rejection\textsuperscript{$\dagger$} & 297 & 0.23 & 0.23 & 0.96 \\
AL-O Organ & 29 & 0.91 & 0.91 & 0.93 \\
AL-P Part / Segment & 76 & 0.64 & 0.64 & 0.71 \\
AL-W Wall & 35 & 0.51 & 0.50 & 0.61 \\
MA-T Shape \& Type & 157 & 0.68 & 0.68 & 0.73 \\
MA-S Surface & 83 & 0.58 & 0.58 & 0.64 \\
MA-C Color & 45 & 0.59 & 0.59 & 0.68 \\
MA-M Margins & 38 & 0.78 & 0.77 & 0.85 \\
MA-SZ Size & 74 & 0.57 & 0.57 & 0.68 \\
PF-D Distribution \& Extent & 43 & 0.49 & 0.50 & 0.65 \\
PF-M Mucosal Features & 124 & 0.65 & 0.65 & 0.70 \\
PF-U Ulcer / Erosion Features & 33 & 0.70 & 0.70 & 0.76 \\
PF-S Tissue Features & 105 & 0.74 & 0.74 & 0.78 \\
PF-B Contents \& Base & 78 & 0.79 & 0.79 & 0.84 \\
QN-L Lesion Count & 63 & 0.83 & 0.83 & 0.86 \\
QN-S Structure Count & 13 & 0.53 & 0.52 & 0.63 \\
GS-MC Morphological Classification & 50 & 0.67 & 0.68 & 0.80 \\
GS-SS Severity Scoring & 50 & 0.70 & 0.70 & 0.76 \\
GS-EC Extent Classification & 13 & 0.59 & 0.59 & 0.69 \\
GS-AG Anatomical Grading & 37 & 0.45 & 0.44 & 0.60 \\
CF-B Bleeding & 69 & 0.96 & 0.96 & 0.97 \\
CF-S Stricture \& Obstruction & 63 & 0.69 & 0.69 & 0.74 \\
CF-A Associated Findings & 37 & 0.60 & 0.60 & 0.64 \\
CR-E Etiology \& Context & 61 & 0.49 & 0.49 & 0.58 \\
CR-I Impression \& Activity & 51 & 0.70 & 0.68 & 0.73 \\
AR-B Biopsy & 84 & 0.91 & 0.91 & 0.95 \\
\bottomrule
\end{tabular}
\\[2pt]
\begin{minipage}{\textwidth}
\footnotesize
$\dagger$~DI-C and DI-R are near-constant binaries (91.8\% and 97.6\% of
answers are the same option), and Kappa $= (P_o - P_e)/(1 - P_e)$,
where chance agreement $P_e = p^2 + (1-p)^2$ for a binary whose
majority answer has share $p$. As $p \to 1$, $P_e \to 1$ and the
denominator vanishes, so $\alpha$ and Kappa are low although raw
agreement is 88.0\% and 96.4\%. For DI-R, $P_e = 0.953$ leaves only
0.047 above chance, of which the raters capture 0.011, giving Kappa
$= 0.23$. Raw agreement is the informative figure for these two
sub-tasks.
\end{minipage}
\end{table}

\subsection{Model accuracy against the consensus label}
\label{appx:consensus-models}

\begin{table}[h]
\caption{Strict accuracy (\%) on the 2{,}044 consensus-labelled questions,
scored against the recorded EHR label and against the consensus;
$\Delta$ = consensus $-$ recorded (pp). Sorted by consensus accuracy.
Values differ from Table~\ref{tab:leaderboard} because the subset is
not the full benchmark; the quantity compared is the within-model
$\Delta$. ColonR1 is eligible for 79 of the 100 cases (1{,}668
questions), is scored on that subset and is not rank-comparable with
the other 13 models.}
\label{tab:consensus-model}
\centering
\small
\begin{tabular}{lrrrr}
\toprule
Model & Questions & Recorded label & Consensus & $\Delta$ \\
\midrule
Qwen3.6-27B & 2{,}044 & 62.9 & 61.4 & $-$1.4 \\
Qwen3.5-9B & 2{,}044 & 59.9 & 59.5 & $-$0.3 \\
Qwen3-VL-32B & 2{,}044 & 59.4 & 58.3 & $-$1.1 \\
Qwen3-VL-8B & 2{,}044 & 57.3 & 55.9 & $-$1.5 \\
Qwen3.5-27B & 2{,}044 & 55.9 & 55.4 & $-$0.5 \\
Hulu-Med-14B & 2{,}044 & 56.0 & 55.1 & $-$0.9 \\
Hulu-Med-32B & 2{,}044 & 56.2 & 55.0 & $-$1.1 \\
Qwen3.5-4B & 2{,}044 & 53.4 & 52.9 & $-$0.5 \\
Hulu-Med-4B & 2{,}044 & 52.0 & 52.2 & +0.1 \\
Hulu-Med-7B & 2{,}044 & 53.4 & 52.0 & $-$1.4 \\
MedGemma-27B & 2{,}044 & 51.3 & 51.3 & +0.0 \\
MedGemma-4B & 2{,}044 & 48.0 & 47.4 & $-$0.6 \\
MedGemma-1.5-4B & 2{,}044 & 44.5 & 44.0 & $-$0.5 \\
\midrule
\emph{ColonR1} & \emph{1{,}668} & \emph{48.0} & \emph{47.0} & \emph{$-$1.0} \\
\bottomrule
\end{tabular}
\end{table}

Over the 13 models scored on all 2{,}044 questions, the two orderings
correlate at Spearman $\rho$ = 0.96, with 73 of 78 model pairs
concordant, and all five discordant pairs lie within the 52--56\%
band.

%% file: appendix_D_extended_related_work.tex
\section{Extended Related Work}
\label{appx:related-work}

This appendix expands on \S\ref{sec:related} with the broader
related-work landscape that was compressed in the main text. It covers
(D.1) endoscopy multimodal benchmarks beyond the comparators surfaced
in main-text Table~\ref{tab:endo-bench-compare-compact}, (D.2)
endoscopy video corpora and foundation models, (D.3) endoscopy
report-generation and agent systems, and (D.4) multi-turn and
EHR-grounded medical VQA precedents in text-only dialogue, radiology
EHR-VQA and multi-turn radiology vision-language models.

\subsection{Endoscopy multimodal benchmarks}
\label{appx:related-endo-bench}

The closest peers are recent benchmarks that evaluate vision-language
and multimodal models on endoscopy data. Where a benchmark
distinguishes a \emph{training corpus} from an \emph{evaluation
subset}, we report the evaluation subset (the comparable axis) and
note the corpus separately.

\begin{table}[h]
\caption{EndoSCOP-V vs. peer multimodal endoscopy benchmarks
(Y $=$ Yes, N $=$ No). This is the full version of the table
summarised in main-text
Table~\ref{tab:endo-bench-compare-compact}.}
\label{tab:endo-bench-compare-full}
\centering
\resizebox{\linewidth}{!}{%
\setlength{\tabcolsep}{3pt}
\footnotesize
\begin{tabular}{lllllllcc}
\toprule
Benchmark & Eval items & Modality & Procedures & Categories & Tasks & Multi-turn & Multi-disease & EHR-GT \\
\midrule
EndoBench~\cite{liu2025endobench}                  & 6{,}832 VQA            & Image          & EGD $+$ Colon $+$ Capsule $+$ Surgical & 4 scenarios       & 12 $+$ 12 sub & N & N & N \\
GI-Bench~\cite{zhu2026gi}                          & 300-Q Human-AI subset$^{\dagger}$ & Image          & EGD $+$ Colon                       & 20 lesions        & 5 stages    & N & N & N \\
ColonVQA~\cite{ji2025colon}                        & 4{,}568 VQA              & Image          & Colon only                          & 76 findings       & 16 of 18    & N & N & N \\
ColonGPT~\cite{ji2026frontiers}                    & 450K dialogues$^{\ddagger}$ & Image          & Colon only                          & 62 sub-categories & 4 (CLS/DET/SEG/VQA) & N & N & N \\
Colon-Bench~\cite{hamdi2026colon}                  & 528 videos / 464K frames & Video          & Colon only                          & 14 lesions        & 3 (VQA/CLS/OV-VOS) & N & N & N \\
PanEndoX~\cite{johri2025clinician}                 & 420K imgs / 30 datasets  & Image          & EGD $+$ Colon $+$ Capsule $+$ Small-bowel & \textbf{111 GI conditions} & 10 (CLS / grading / landmark) & N & N & N \\
HyperKvasir~\cite{borgli2020hyperkvasir}           & 110{,}079 imgs $+$ 374 vids & Image $+$ video & EGD $+$ Colon                       & 23 classes        & (corpus)    & N & N & N \\
Kvasir-VQA~\cite{gautam2024kvasir}                 & 6{,}500 imgs / ${\sim}$58{,}849 QA & Image          & EGD $+$ Colon                       & 5 categories      & 6 Q-types   & N & N & N \\
Kvasir-VQA-x1~\cite{gautam2025kvasir}              & 159K QA (perturbations) & Image          & EGD $+$ Colon                       & 5 categories      & 3 reasoning levels & N & N & N \\
SurgViVQA~\cite{drago2025surgvivqa}                & 5{,}200 clip-QA & 60 Videos      & Colon only      & Polyps (REAL-Colon source) & 6 domains (diagnosis 3\%) & N & N & N \\
EndoFM-LV~\cite{wang2025improving}                 & 6{,}469 long videos / 13M frames & Long video    & EGD $+$ Colon                       & (FM corpus)       & (FM)        & n/a & n/a & n/a \\
\textbf{EndoSCOP-V (ours)}                         & \textbf{20{,}274 multi-turn QAs / 1{,}307 clips} & \textbf{Video clips} & \textbf{EGD (1{,}004) $+$ Colon (303)} & \textbf{54 diseases} & \textbf{9 $\times$ 30 sub} & \textbf{Y} & \textbf{Y (22\%)} & \textbf{Y} \\
\bottomrule
\end{tabular}}

\smallskip
\footnotesize
$^{\dagger}$ GI-Bench: The authors describe a 5-questions-per-image
protocol with model-vs-human comparison on a 60-image (300-question)
stratified subset.
$^{\ddagger}$ ColonINST's ``human-machine dialogues'' are
LLM-generated instruction-tuning samples, not clinical multi-turn
consultations. The underlying task framing remains single-turn image
VQA.
\end{table}

\paragraph{EndoBench}~\cite{liu2025endobench} is the closest direct image-VQA peer. It
evaluates 23 MLLMs on a 6{,}832-item validated subset distilled from a
446{,}535-pair source corpus (EndoVQA-Instruct), with a 2-clinician
reference standard. EndoBench's additional scenarios (capsule and
surgical endoscopy) are not straightforwardly an ``extra breadth''
axis: capsule is a passive, operator-uncontrolled modality that drifts
through the GI tract and cannot dwell, retroflex, or deploy
instruments, while surgical endoscopy is an operative regime with
different visual content. EGD and colonoscopy are operator-controlled
diagnostic modalities in which the endoscopist actively decides where
to dwell, biopsy, or treat. EndoSCOP-V is scoped to this
diagnostic-reporting regime and does not claim to generalize to
capsule or surgical settings. What we add over EndoBench is video
clips (vs. images), multi-turn dialogue evaluation under
context-pruned oracle-forcing (\S4), multi-disease case mix, and
deterministic EHR-grounded ground truth.

\paragraph{GI-Bench}~\cite{zhu2026gi} organizes 20 fine-grained lesion categories around a
five-stage workflow and benchmarks 12 MLLMs against three junior
endoscopists and three residency trainees. Top-tier proprietary models
(Gemini-3-Pro at Macro-F1 = 0.641) match junior endoscopists (0.727)
on diagnosis but fail on spatial grounding (mIoU 0.345 vs. human
0.506). The authors describe a ``fluency-accuracy paradox'' where MLLM
reports read smoothly but are factually weaker. EndoSCOP-V differs
along the same axes as above, and our clinician baseline also runs on
a substantially larger 1{,}500-question subset (vs. GI-Bench's
300-question Human-AI panel).

\paragraph{Colon-X / ColonVQA, ColonINST, ColonGPT, and the broader ``Frontiers in Intelligent Colonoscopy'' survey}~\cite{ji2025colon,ji2026frontiers} together constitute the largest
data-and-model effort in colonoscopy. ColonVQA provides 1.1M
training-corpus VQA pairs over 212K images and 76 findings, and
\textbf{ColonEval} is the actual benchmark subset (4{,}568 VQA over 16
of the 18 tasks). ColonINST, the companion instruction-tuning dataset,
contains 303{,}001 colonoscopy images across 62 sub-categories from 19
public sources, with 128{,}620 GPT-4V-generated medical captions and
450{,}724 human-machine dialogues, but those ``dialogues'' are
instruction-tuning samples (single-image / single-turn
instruct$\rightarrow$answer pairs in a chat-format wrapper), not
clinical multi-turn consultations. The Frontiers survey investigates
63 colonoscopy datasets and 137 deep models. Colon-X is
\textbf{colonoscopy-only}, \textbf{image-based}, and
\textbf{single-turn} at evaluation time. We evaluate ColonR1 (the
reasoning-fine-tuned model from this group) on EndoSCOP-V (\S5) as one
of our 14 open-weight models. Its performance on our upper-GI clips
and multi-turn protocol tests how far a colonoscopy specialist
generalizes outside its training distribution.

\paragraph{Colon-Bench}~\cite{hamdi2026colon} uses a multi-stage
agentic workflow to densely annotate 528 full-procedure colonoscopy
videos with 14 lesion categories (464{,}035 frames, 300{,}132
bounding boxes, 213{,}067 segmentation masks, 133K words of clinical
descriptions), evaluating lesion classification, Open-Vocabulary
Video Object Segmentation, and single-turn VQA. Colon-Bench is
\textbf{colonoscopy-only}, \textbf{single-turn}, has no
multi-disease case structure, and uses agent-generated descriptions
clinician-reviewed for quality rather than EHR-grounded ground truth.

\paragraph{SurgViVQA}~\cite{drago2025surgvivqa} is a \textbf{surgical VideoQA framework},
the closest published \emph{video-grounded} peer for endoscopy VQA,
built on a tube-masked Video--Text Encoder architecture. The framework
evaluates jointly on \textbf{two datasets}: their newly released
\textbf{REAL-Colon-VQA} (5{,}200 clip-QA pairs, 4{,}450 train / 750
test, curated from REAL-Colon's \textbf{60 polyp-annotated colonoscopy
procedures}~\cite{biffi2024real}) and the publicly available
\textbf{EndoVis18-VQA} (13{,}790 QA from 2{,}086 surgical scenes
across 14 nephrectomy videos). Endoscopy reporting is not the
framework's primary target. The authors claim REAL-Colon-VQA as the
first publicly available colonoscopy video Q\&A dataset, a claim
contemporaneous with Colon-Bench (above), which evaluates single-turn
VQA on full-procedure colonoscopy videos. REAL-Colon-VQA is
\textbf{polyp-centric} by inheritance, since REAL-Colon's underlying
350K bounding-box and histopathology annotations are for polyps and
normals, and its question mix reflects this. The 20 question
categories are dominated by Movement (scope / lesion motion, 31.73\%)
and Operation Notes (illumination, visibility, flushing, occlusion,
46.15\%), while Diagnosis (histopathology) is only 2.88\% and Sizing
2.88\%. The bulk of QA therefore targets short-horizon endoscopic
dynamics rather than per-disease clinical reasoning. Methodologically,
the SurgViVQA model processes \textbf{8~frames sampled every 4~frames
from REAL-Colon's 30 fps stream, a temporal span of
${\approx}$0.93~s}. This deliberate input-window choice targets
short-horizon dynamics observable within ${\sim}$1~s and is the
model's input window, not a dataset-wide microclip property.
EndoSCOP-V differs in (a) covering EGD as well as colonoscopy, (b)
spanning \textbf{54 diseases + 4 non-disease findings} rather than a
single polyp-focused disease family, (c) drawing on \textbf{1{,}307
cases} vs. 60, (d) presenting clinical-clip-length \textbf{HD video}
(mean ${\approx}$174~s, \textbf{97\% 1080p}, with per-disease
key-frame anchors), (e) being multi-turn with explicit oracle-forcing
rather than single-turn and (f) deriving ground truth
deterministically from clinician-reviewed EHR templates rather than
from polyp annotations augmented with motion- and visibility-attribute
Q\&A pairs.

\paragraph{PanEndoSuite / PanEndoX / PanEndoFM}~\cite{johri2025clinician} is the nearest concurrent benchmarking
effort built on the same WEO MST-3.0 terminology backbone we use.
PanEndoAtlas harmonises 30 public endoscopy datasets across 13
countries and 26 hospitals into 420{,}000 labeled images organised
under MST-3.0. PanEndoX defines 10 image-level tasks (hierarchical
GI-tree classification, Barrett's grading, UC scoring, polyp
subtyping, BBPS, anatomical-landmark ID, multi-organ disease
classification), and PanEndoFM is the companion 10M-image DINOv2
foundation model benchmarked against EndoFM-LV, EndoSSL, ViT-B/16, and
ResNet-50. PanEndoX and EndoSCOP-V occupy \textbf{fundamentally
different axes} of the endoscopy-evaluation problem. PanEndoX is
\textbf{image-only}, \textbf{single-turn}, and \textbf{closed-set
classification/grading}, with \emph{one task per image} and a broad
label space. EndoSCOP-V is \textbf{video-grounded},
\textbf{multi-turn}, and \textbf{deeply per-disease}. Across the 58
findings in our taxonomy (54 strict diseases + 4 non-disease
landmarks/findings), the benchmark carries \textbf{1{,}033 distinct,
clinician-reviewed question templates, an average of ${\sim}$18 per
disease}, spanning the \textbf{9 tasks $\times$ 30 sub-tasks} of the
clinical reporting workflow (detection, anatomical localisation,
measurement, multi-variable grading, intervention reasoning, report
drafting). These templates instantiate per-case against the EHR,
producing the \textbf{20{,}274 multi-turn questions over 1{,}307
cases} that constitute the full evaluation surface. Where PanEndoX
asks ``is this image disease X?'' once, EndoSCOP-V asks the
operator-level chain of follow-up questions (location, size,
morphology, severity grade, biopsy / treatment recommendation, report
drafting) that turns a single positive finding into an actual clinical
episode. The two benchmarks therefore answer different questions about
a model: PanEndoX measures \emph{image-level recognition} across the
GI label space, whereas EndoSCOP-V measures \emph{per-disease clinical
reasoning depth} over the reporting workflow. A model evaluated on
both is characterised much more fully than by either alone.

\paragraph{HyperKvasir}~\cite{borgli2020hyperkvasir} is the foundational public GI dataset
(110{,}079 images + 374 videos / 23 classes / both upper + lower GI /
B\ae{}rum Hospital, Norway). It is a corpus rather than a benchmark,
with image-level class labels rather than multi-turn QA, but it
underpins much subsequent work.
\textbf{Kvasir-VQA}~\cite{gautam2024kvasir} extended the SimulaMet
line into VQA with 6{,}500 expertly-curated images (${\sim}$58{,}849
QA pairs) across 5 image categories (Normal, Polyps, Esophagitis,
Ulcerative Colitis, Instrument) and six question types (yes/no,
single-choice, multi-choice, color, location, count).
\textbf{Kvasir-VQA-x1}~\cite{gautam2025kvasir} adds 159{,}549
LLM-generated QA pairs stratified by reasoning complexity (1--3) plus
image-level perturbations (occlusion, contrast, blur) for robustness
testing. The Kvasir-VQA family is the closest sustained image-VQA
effort covering both upper and lower GI but is \textbf{single-turn}
with QA generated by experts/LLMs from class labels rather than from
clinical records.

EndoSCOP-V's structural novelties relative to this group are: (a)
\textbf{multi-turn video evaluation} at clinical scale, (b)
\textbf{multi-disease case mix} with up to six co-existing diagnoses
per case, (c) the \textbf{Markovian context-pruned oracle-forcing
protocol} (\S4) that grades long clinical dialogues without cascading
dialogue errors, and (d) \textbf{deterministic EHR-grounded ground
truth} rather than caption-paired or LLM-generated labels.

\subsection{Endoscopy video corpora and foundation models}
\label{appx:related-corpora-fm}

A separate line of work provides large endoscopy video datasets as
\emph{training corpora} or foundation-model pre-training resources
rather than VLM benchmarks. \textbf{REAL-Colon}~\cite{biffi2024real}
compiles 60 full-resolution real-world colonoscopy videos from six
medical centers (2{,}757{,}723 native $1920\times1080$ frames) with
350K bounding-box polyp annotations, plus patient clinical data and
polyp histopathology, a unique full-procedure resource but limited to
a single pathology family. \textbf{Galar}~\cite{le2025galar} provides
80 video capsule endoscopy videos from two German centers with
3{,}513{,}539 framewise-annotated frames across 29 labels, the largest
open VCE dataset to date but in a fundamentally different acquisition
regime, because the capsule drifts passively through the GI tract and
cannot dwell on suspicious areas, retroflex, or deploy tools. The
visual distribution and clinical workflow are not interchangeable with
operator-controlled diagnostic endoscopy.
\textbf{EndoFM-LV}~\cite{wang2025improving} curates 6{,}469 long
endoscopy videos for foundation-model pre-training without
question-level evaluation.

A parallel line develops self- or weakly-supervised foundation models
for endoscopy frames and short clips.
\textbf{EndoFM}~\cite{wang2023foundation} pre-trains a 121M-param
video transformer on 33K clips (${\sim}$5M frames).
\textbf{EndoSSL}~\cite{hirsch2023self} builds Masked Siamese Networks
on 7{,}700 laparoscopy + 14{,}000 colonoscopy videos.
\textbf{EndoDINO}~\cite{dermyer2025endodino} scales DINOv2 to ViT-1B
on 3.5B raw frames sampled from 130{,}037 source videos.
\textbf{GastroNet-5M}~\cite{boers2024foundation} presents a 5M-image
SSL corpus from eight Dutch hospitals. \textbf{A Foundational
Multi-Task Multimodal Model for UGI
Endoscopy}~\cite{he2025foundational} is the closest multimodal
upper-GI foundation model, framing four downstream tasks under
instruction-tuning. None of these papers releases a benchmark for
general-purpose VLM evaluation, since they target encoder quality via
narrow downstream tasks. EndoSCOP-V is complementary, a multi-turn,
disease-broad, clinically-anchored evaluation surface that any of the
above corpora's pre-trained encoders can be tested against.

\paragraph{Polyp segmentation and multi-class image classification} form a deep, separate lineage of endoscopy datasets that pre-date the
VQA era. \textbf{SUN-SEG}~\cite{ji2022video} densely annotates
158{,}690~frames across 1{,}013 colonoscopy video sequences with five
label types (object mask, boundary, scribble, polygon, attribute) and
benchmarks 13 polyp/object segmentation methods.
\textbf{GastroVision}~\cite{jha2023gastrovision} takes the multi-class
image classification angle, releasing 8{,}000 images across 27
anatomical-landmark, pathological-finding, polyp-removal, and normal
classes from two hospitals (B\ae{}rum, Norway and Karolinska, Sweden).
These are single-turn, image-level, classification/segmentation
benchmarks that constitute the deep-skill layer that EndoSCOP-V's
task-level analysis (\S5) draws on but does not subsume.

\paragraph{Temporal segmentation and procedure-specific safety benchmarks} form a third complementary track.
\textbf{ColonTCN}~\cite{biffi2025temporal} re-annotates the REAL-Colon
corpus (60 videos, 2.7M frames) with frame-level labels across 9
anatomical / procedural-phase categories, providing the first
open-access benchmark for full-procedure colonoscopy temporal
segmentation and a custom temporal-convolutional baseline.
\textbf{BleedOrigin-Bench}~\cite{xu2025bleedorigin} addresses ESD
intra-operative safety with 1{,}771 expert-annotated bleeding sources
across 106{,}222~frames from 44 endoscopic-submucosal-dissection
procedures (8 anatomical sites, 6 clinical scenarios).
\textbf{TEMSET-24K}~\cite{bilal2025temset} densely annotates 24{,}306
trans-anal-endoscopic-microsurgery (TEMS) microclips with hierarchical
\emph{phase / task / action} triplets (5 / 12 / 21). These benchmarks
operate at the \emph{procedure-step} level (where in time a phase,
action or bleeding source is) rather than at the
\emph{clinical-finding-reporting} level. EndoSCOP-V's per-disease
key-frame anchors give a related but coarser temporal hook (which
frame in the clip best supports a finding) for each diagnosis, which
is the temporal granularity required for VLM reporting evaluation,
while fine-grained phase segmentation (ColonTCN, TEMSET-24K) and
sub-procedural safety detection (BleedOrigin-Bench) remain orthogonal
problems.

\paragraph{Adjacent surgical multimodal evaluation.} \textbf{SurgVLM}~\cite{zeng2025surgvlm} is the most comprehensive
surgical VLM benchmark to date. Its SurgVLM-DB unifies 23 public
surgical datasets into 1.81M frames and 7.79M conversations across 16
surgical types and 18 anatomical structures, and SurgVLM-Bench
evaluates 14 mainstream commercial VLMs (GPT-4o, Gemini 2.0 Flash,
Qwen2.5-Max inter alia) over 10 surgical tasks ranging from instrument
recognition to critical-view-of-safety reasoning. SurgVLM and
EndoSCOP-V are nearest analogues \emph{across procedural domains},
since SurgVLM is to surgical/laparoscopic intelligence what EndoSCOP-V
is to diagnostic GI endoscopy reporting. The two domains differ in
clinical question structure (intra-operative tool/anatomy/phase vs.
diagnostic finding/grading/recommendation), in primary modality
(operative camera vs. diagnostic endoscopy), and in evaluation form
(SurgVLM is single-turn instruction-following over agglomerated public
datasets, whereas EndoSCOP-V is multi-turn dialogue over a single
case-grounded EHR-anchored corpus).

\subsection{Endoscopy report generation and agent systems}
\label{appx:related-reporting-agents}

A growing line of work develops models that \emph{generate} endoscopy
outputs, such as full reports or sequenced agentic decisions, rather
than answer isolated questions.

\paragraph{From Scope to Script}~\cite{kaklamanos2025scope} proposes a two-stage transformer
encoder--decoder for GI report generation, pre-trained on 775{,}990
image-text pairs from 32{,}949 EGDs and 72{,}314 colonoscopies
(78{,}537 patients). \textbf{Report-Angel}~\cite{jiang2026domain} is a
domain-specific MLLM trained on 20{,}617 image-text pairs and
prospectively validated across multiple centers, achieving a
clinically-acceptable report rate of \textbf{79.3\% (internal) /
83.3\% (external)}, case-level report completeness 88.51\%, accuracy
78.93\%, and lesion-level reporting accuracies of 91.92\%
(retrospective image), 89.07\% (single-center prospective video), and
83.94\% (multi-center prospective video). \textbf{Massimi et
al.}~\cite{massimi2026large} conduct a within-case paired pilot of
Gemini 2.5 Pro on 5 EGD videos with and without an EndoAngel CAD
overlay, rated by 5 blinded endoscopists. They find landmark-agreement
accuracy of 0.55 (clean-video) vs. 0.33 (overlay-video) and conclude
that \emph{``Gemini 2.5 Pro demonstrated inadequate performance for
clinical EGD reporting''}, a hypothesis-generating result that
motivates the kind of standardized, turn-by-turn evaluation surface we
provide. The 5 EGD videos are not publicly released, and the paper's
anonymisation note (\emph{``All videos were anonymized before
analysis''}) refers to cloud-API processing rather than a public
corpus release.

\paragraph{EndoAgent}~\cite{tang2025endoagent} is the first memory-guided reflective agent
for endoscopic vision-to-decision reasoning. EndoAgent uses a
dual-memory architecture (short-term memory for action traces,
long-term memory for reflective summaries of errors and uncertainties)
to coordinate a toolset spanning six core tasks (lesion
classification, detection, segmentation, image editing, VQA, and
medical report generation). It is evaluated on
\textbf{EndoAgentBench}, a benchmark of \textbf{5{,}709
expert-annotated visual-QA pairs} across five diagnostic subtasks.
EndoAgent and EndoSCOP-V occupy adjacent but complementary positions.
EndoAgent benchmarks an \emph{agentic system} that orchestrates
external tools and memory, while EndoSCOP-V evaluates the
\emph{intrinsic multi-turn reasoning} of off-the-shelf LMMs without
tool calls, and EndoAgentBench is image-grounded with 5{,}709 queries,
whereas EndoSCOP-V is video-grounded with 20{,}274 turns. Future work
could naturally pair the two by running EndoAgent on EndoSCOP-V to
test how much agentic tool use closes the 23-percentage-point gap
(\S5) we observe between the best non-agentic frontier model and the
clinician panel on the between-subjects spot-check.

These reporting and agent systems are \emph{generative endpoints}
with locally-curated rater panels or per-system rubrics. EndoSCOP-V
is positioned to be the standardized external evaluation surface
that they can be tested against, with disease-broad coverage, a
multi-turn protocol that decomposes the reporting workflow into 9
tasks $\times$ 30 sub-tasks, and a clinician baseline (78\% accuracy on 1{,}500 questions) and a
five-rater consensus re-annotation (2{,}148 questions) to anchor
accuracy claims.

\subsection{Multi-turn and EHR-grounded medical VQA}
\label{appx:related-multi-turn}

Multi-turn evaluation of LLMs and multimodal models in clinical
settings has matured rapidly along three threads: text-only multi-turn
medical dialogue, EHR-grounded VQA in radiology, and multi-turn
radiology vision-language models. Each is well developed, but none has
yet been combined with endoscopy video at multi-turn scale.

\begin{table}[h]
\caption{Multi-turn and EHR-grounded medical VQA / dialogue
benchmarks. \emph{Vis} $=$ vision-grounded.}
\label{tab:multi-turn-bench}
\centering
\resizebox{\linewidth}{!}{%
\setlength{\tabcolsep}{3pt}
\footnotesize
\begin{tabular}{lllccl}
\toprule
Benchmark & Modality / Source & Items & Multi-turn & Vis & Domain \\
\midrule
emrQA~\cite{pampari2018emrqa}                       & Text                         & 455{,}837 QA / 1.3M logical-form pairs / 2{,}425 notes & partly$^{\dagger}$ & N & EHR \\
MedDialog~\cite{zeng2020meddialog}                  & Text                         & 3.4M Chinese / 0.26M English consultations & Y & N & General medicine \\
Rad-ReStruct / hi-VQA~\cite{pellegrini2023rad}      & CXR                          & hierarchical structured-report & \textbf{Y, hierarchical} & Y & Radiology \\
MIMIC-Diff-VQA~\cite{hu2023medicaldiff}             & CXR (MIMIC-CXR)              & 700k QA / 164k image pairs & longitudinal pairs & Y & Radiology \\
EHRXQA / MIMIC-CXR-VQA~\cite{bae2023ehrxqa}         & CXR $+$ structured EHR       & image$+$table QA & partly & Y & Radiology $+$ EHR \\
RaDialog~\cite{pellegrini2025radialog}              & CXR                          & dialog-instruct & \textbf{Y} & Y & Radiology \\
RadVLM~\cite{deperrois2025radvlm}                   & CXR                          & 1M$+$ image-instruction pairs (single $+$ multi-turn) & \textbf{Y} & Y & Radiology \\
HealthBench~\cite{arora2025healthbench}             & Text                         & 5{,}000 multi-turn convs / 48{,}562 rubrics & Y & N & Gen.\ medicine \\
MedDialogRubrics~\cite{gong2026meddialogrubrics}    & Text (synthesized)           & 5{,}200 cases / 60K EBM rubrics / up to 13 turns & Y & N & Gen.\ medicine \\
MedMT-Bench~\cite{yang2026medmt}                    & Text (synthesized)           & 400 long convs / avg 22 turns & Y & N & Diagnosis $+$ treatment \\
ThReadMed-QA~\cite{munnangi2026threadmed}           & Text                         & 2{,}437 convs / 8{,}204 QA / up to 9 turns & Y & N & Health Q\&A \\
Gastric-X~\cite{lu2026gastric}                      & 3D CT $+$ endoscopic image   & 1.7K cases (5 tasks) & N & Y & Gastric cancer (CT-led) \\
\textbf{EndoSCOP-V (ours)}                          & \textbf{Video (EGD $+$ colonoscopy)} & \textbf{20{,}274 turns / 1{,}307 cases} & \textbf{Y, oracle-forced} & \textbf{Y} & \textbf{Endoscopy reporting} \\
\bottomrule
\end{tabular}}

$^{\dagger}$ emrQA's questions are templated and
mainly single-turn given a note, and multi-turn structure emerges from
longitudinal patient records but is not the explicit evaluation axis.
\end{table}

\paragraph{Text-only medical
dialogue.}~\textbf{MedDialog}~\cite{zeng2020meddialog} is the largest medical
dialogue corpus to date (3.4M Chinese patient-doctor consultations
covering 172 specialties and 0.26M English covering 96 specialties),
drawn from haodf.com, healthcaremagic.com, and icliniq.com.
\textbf{MedDG}~\cite{liu2022meddg} adds an entity-centric Chinese
consultation set. The most recent generation operationalizes
multi-turn evaluation through synthesized cases and rubric-based
grading: \textbf{HealthBench}~\cite{arora2025healthbench} reports
5{,}000 multi-turn conversations evaluated against 48{,}562 rubric
items validated by 262 physicians,
\textbf{MedDialogRubrics}~\cite{gong2026meddialogrubrics} constructs
5{,}200 patient cases with over 60{,}000 fine-grained
Evidence-Based-Medicine rubrics across up to 13 consultation turns,
and \textbf{MedMT-Bench}~\cite{yang2026medmt} builds 400 long medical
conversations averaging 22 rounds (max 52) and operationalizes five
challenge dimensions (long-context memory, contextual interference,
self-correction, instruction clarification, multi-instruction
response). All 17 frontier LLMs evaluated on MedMT-Bench score below
60\%, with the best at 59.75\%.
\textbf{ThReadMed-QA}~\cite{munnangi2026threadmed} takes the
alternative route of authentic patient threads, mining r/AskDocs for
2{,}437 patient-physician conversations (8{,}204 QA pairs across up to
9 turns) and finds that even GPT-5 achieves only 41.2\% fully-correct
responses, with wrong-answer rates roughly tripling by turn three.
They introduce per-turn metrics (Turn-Level Degradation,
Conversational Consistency Score, Error Propagation Rate) that
quantify cascading dialogue errors, precisely the failure mode our
oracle-forcing protocol (\S4) controls for.

\paragraph{EHR-grounded VQA in
radiology.}~\textbf{emrQA}~\cite{pampari2018emrqa} established the canonical
text-only EHR-QA dataset by re-purposing i2b2 clinical-note
annotations into 1M question-logical-form pairs and 400K+ QA evidence
pairs over five categories (relations, medications, heart disease,
obesity, smoking). \textbf{EHRXQA}~\cite{bae2023ehrxqa} is the first
multi-modal EHR-QA dataset combining structured EHRs (MIMIC-IV) with
chest X-ray imaging (MIMIC-CXR + Chest ImaGenome scene graphs), with
three QA scopes (image-related, table-related, image+table-related).
Its companion \textbf{MIMIC-CXR-VQA} provides image-only EHR-QA over
chest radiographs with 48 templates $\times$ 7 question types.
\textbf{MIMIC-Diff-VQA}~\cite{hu2023medicaldiff} introduces 700{,}703
QA pairs over 164{,}324 main--reference chest-X-ray image
\emph{pairs}, organized by the clinical
Assessment-Diagnosis-Intervention-Evaluation procedure to capture
longitudinal reasoning. These benchmarks establish that EHR-grounded
VQA is well developed in radiology, but every endoscopy VQA benchmark
surveyed (Kvasir-VQA, Kvasir-VQA-x1, ColonINST, EndoBench,
EndoVQA-Instruct, ColonVQA, GI-Bench, Colon-Bench) generates questions
from image captions, label templates or LLM prompting on existing
class labels, never from clinical pathology reports or endoscopy
notes. \textbf{Gastric-X}~\cite{lu2026gastric} is the closest
gastroenterology-domain analogue of the radiology multi-modal /
EHR-grounded line. It holds \textbf{1.7K gastric-cancer-only cases}
(single-disease scope), each pairing one endoscopic still
(${\sim}$1{,}700 endoscopy images total) with multi-phase 3D CT
(non-contrast, arterial, venous, equilibrium), structured biochemistry
(WBC, neutrophil, serum tumour markers), expert-authored diagnostic
reports, and 3D bounding boxes for tumour and perigastric lesions, and
its five tasks include VQA, report generation, cross-modal retrieval,
classification, and lesion localisation. Gastric-X grounds VQA in
structured labs and clinician reports, a principle our protocol
shares, but its primary visual signal is \textbf{CT volume} with
endoscopy as a side channel, and the protocol is single-turn.
EndoSCOP-V is the corresponding multi-turn, endoscopy-video-led peer
to that multimodal-clinical-grounding tradition.

\paragraph{Multi-turn radiology vision-language models (the
methodological precedent).}~\textbf{Rad-ReStruct / hi-VQA}~\cite{pellegrini2023rad} is the closest
published methodological precedent for our oracle-forcing protocol.
The authors model structured radiology reporting as
\textbf{hierarchical visual question answering} and propose hi-VQA,
\emph{``a hierarchical, autoregressive VQA method\ldots [that]
considers the prior context of previously asked questions and answers,
allowing to exploit inter-dependencies between questions about the
same image''}. Their hierarchy is three levels (Topic Existence
$\rightarrow$ Element Existence $\rightarrow$ Attributes) over
chest-X-ray structured reports, populated autoregressively by carrying
earlier turns forward as context, precisely the
\emph{forced-context-at-next-turn} mechanic we adopt and generalize.
\textbf{RaDialog}~\cite{pellegrini2025radialog} (same TUM CAMP group
as Rad-ReStruct) extends this line to a full conversational radiology
assistant with a dual-branch structured-findings architecture,
achieving 13.3\% improvement in clinical correctness and being
preferred by radiologists in 84\% of cases over comparators.
\textbf{RadVLM}~\cite{deperrois2025radvlm} is the most recent
multitask conversational CXR foundation model, curated on 1M+
image-instruction pairs that explicitly include both single-turn and
multi-turn conversational interactions generated by GPT-4o-mini from
CXR reports, abnormality labels, and bounding boxes.

\paragraph{What EndoSCOP-V extends.} Our Markovian context-pruned
oracle-forcing protocol (\S4) builds directly on the hi-VQA family's
idea of carrying prior Q\&A turns forward as context, but extends it
from a 3-level hierarchy on chest-X-ray structured reports to a
\textbf{9 $\times$ 30 sub-task taxonomy over multi-turn endoscopy
video dialogue}, replaces \emph{autoregressive prediction
propagation} with \textbf{explicit gold-answer forcing} between
turns (so a downstream sub-task is graded on reasoning at
\emph{that} turn rather than on potentially-wrong earlier
predictions), and grounds the question taxonomy in
clinician-reviewed EHR templates rather than radiology-report text
mining. The cascading-dialogue-error phenomena documented
quantitatively by ThReadMed-QA (turn-level degradation, error
propagation rate) are exactly what oracle forcing is designed to
control for in the endoscopy setting.